\documentclass[final,5p,times,twocolumn]{elsarticle}

\usepackage{amssymb}

\usepackage{amsmath,amsfonts}
\usepackage{bm}
\usepackage{nomencl}
\makenomenclature
\usepackage{algorithm}
\usepackage{algpseudocode}
\usepackage{array}
\usepackage{subfig}
\usepackage{color}
\usepackage{xcolor}
\newcolumntype{M}[1]{>{\centering\arraybackslash}m{#1}}
\usepackage{array}
\usepackage{multicol}
\usepackage{multirow}
\usepackage{url}
\usepackage{graphicx}
\usepackage{bm}
\usepackage{comment}
\usepackage{soul}
\usepackage{hyperref}

\newtheorem{theorem}{Theorem}

\newtheorem{remark}[theorem]{Remark}

\journal{Applied Energy}

\begin{document}

\begin{frontmatter}

\title{Interpretable Kolmogorov-Arnold Network with Feature-Isolated Temporal Attention Mechanism for Electricity Load Forecasting}

\author[label1]{Jinhao Li}
\ead{jinhao.li@monash.edu}
\author[label1,label2]{Hao Wang\corref{cor1}}
\ead{hao.wang2@monash.edu}
\cortext[cor1]{Corresponding author: Hao Wang.}
\affiliation[label1]{
            organization={Department of Data Science and AI, Faculty of Information Technology, Monash University},
            city={Melbourne},
            state={Victoria},
            country={Australia}
}
\affiliation[label2]{
            organization={Monash Energy Institute, Monash University},
            city={Melbourne},
            state={Victoria},
            country={Australia}
}

\begin{abstract}
Accurate electricity load forecasting is a crucial prerequisite for stable power system operations. While prevalent deep learning models present competitive performance, they often operate as black boxes and lack interpretability. However, such capability has increasingly become a key enabler to understand the influences of forecast drivers, thereby assisting more informed system operations. While the Kolmogorov-Arnold network (KAN) has emerged as a promising alternative because of its learnable activation function design, its direct application to time-series forecasting faces challenges in modeling complex temporal data patterns. Also, simple integration into existing architectures, such as serving as replacement of neural modules, cannot fully leverage KAN's interpretability strengths. To address these gaps, this study develops LoadKAN, a novel hybrid and interpretable framework for load forecasting that synergistically combines a specifically-designed feature-isolated temporal attention mechanism with a KAN module. The attention stage aims to extract temporal dynamics from each input feature independently, such as historical load and human mobility, providing distilled feature representations to the KAN module for interpretable predictions. When evaluated on datasets from three representative U.S. electricity markets, our LoadKAN remains highly competitive when compared to extensively-tuned, state-of-the-art, black-box deep learning benchmarks. More importantly, LoadKAN's interpretability enables a granular analysis of the learned non-linear relationships between six distinct mobility patterns and electricity load. Through KAN-learned activation functions, our quantitative sensitivity analyses on mobility features reveal complex and market-specific dependencies. These findings further demonstrate the ability of our LoadKAN to generate insights often obscured by opaque black-box neural forecasting models.
\end{abstract}

\end{frontmatter}

\section{Introduction} \label{sec:intro}
Accurate electricity load forecasting is a cornerstone of modern energy management, which is essential for maintaining the stability, efficiency, and economic viability of power systems worldwide~\cite{Maghraoui2024,Wang2024_intro}. Inaccuracies in load forecasting can lead to significant operational inefficiencies, including suboptimal generation dispatch, increased reliance on expensive peaking power plants, and compromised grid reliability~\cite{Yang2023}, potentially incurring substantial operational costs~\cite{Marshman2022} and, in severe cases, supply disruptions~\cite{Davis2024}. Even modest errors can translate into millions of dollars in unnecessary expenses or lost revenue~\cite{Marshman2022}. The global power generation market, valued at approximately US\$1.67 trillion in 2023~\cite{NextMove2024}, is projected to reach nearly US\$2.9 trillion by 2030, with global electricity demand anticipated to grow by an average of 3.4\% annually through 2026~\cite{IEA2024}. This scale and growth highlight the critical nature of precise demand predictions. Therefore, advancing load forecasting methodologies is essential to navigating the complexities of an evolving energy landscape featured by increasing renewable energy integration~\cite{Rubasinghe2023,Hyunsik2025}, more complex demand patterns from growing adoption of electric vehicles, and dynamic human activities~\cite{wang2024reinforcement}.

\subsection{Motivation} \label{subsec:intro_motivation}
Traditional statistical models, such as autoregressive integrated moving average (ARIMA)~\cite{Kuster2017}, often struggle to capture the inherent non-linearities in time-series data. While contemporary deep learning models -- such as recurrent neural network (RNN)-based~\cite{Xia2021}, convolutional neural network (CNN)-based~\cite{Wang2024}, graph neural network (GNN)-based~\cite{li2025multi,Mansoor2024,Giamarelos2024}, and transformer-based architectures~\cite{Peng2025,Jiang2024_att} -- have demonstrated strong predictive performance, they often operate as ``black boxes'', hindering a deeper understanding of forecast drivers. However, such forecast interpretability has been increasingly recognized as an essential enabler in the field of electricity load forecasting to understand the impact of complex and dynamic exogenous features on electricity consumption. Among these features, human mobility patterns have emerged as significant predictors of electricity demand~\cite{Chen2020, Liu2022, Zarbakhsh2022}, reflecting the collective behavior of consumers and the operational rhythms of commercial and industrial sectors~\cite{Cordova2019, Zhang2023}. However, the precise relationship between diverse mobility streams (e.g., activities in retail areas, workplaces, transit hubs, and residential areas) and electricity load is complex, non-linear, and often varies considerably across different geographical regions and electricity market characteristics~\cite{Prabowo2023, Obst2021}. Unraveling these relationships is therefore crucial for gaining deeper insights into energy consumption dynamics. Consequently, a pressing need exists for forecasting models that achieve high accuracy while also providing clear and interpretable insights into how input features (particularly human mobility) influence electricity load.

The presence of the Kolmogorov-Arnold network (KAN) \cite{kan} offers a compelling alternative paradigm. Unlike most neural networks that employ fixed and often simple activation functions at the neuron nodes, KAN features learnable activation functions, represented as splines, directly on the network's edges. This architectural distinction is key: while black-box neural models may necessitate separate and post-hoc interpretability techniques~\cite{Baur2024}, such as Local Interpretable Model-agnostic Explanations (LIME)~\cite{lime} or Shapley Additive exPlanations (SHAP)~\cite{shap}, to approximate a model's decision-making, KAN is capable of embedding interpretability directly within its structure. Post-hoc methods, though useful, act as external auditors in nature, attempting to approximate a model's complex logic. These approximations can be unstable or even misleading, potentially providing a false sense of understanding while obscuring the true drivers of a forecast~\cite{Slack2021}. In contrast, the learnable splines in KAN explicitly represent the learned functional relationships between individual input variables and the subsequent layer or output, thereby reducing reliance on potentially unreliable post-hoc tools and providing an inherently transparent mechanism for understanding feature contributions.

Inspired by the distinguished neural architecture of KAN,  the principal aims of this study are summarized as follows.
\begin{itemize}

    \item Explore KAN's applicability for electricity load forecasting, in particular its inherent architectural transparency in elucidating how input features drive forecast outcomes, along with evaluating the predictive accuracy of KAN-based models against established deep learning benchmarks to ensure viability.
    
    \item Leverage KAN's interpretability via its learnable spline-based activation functions to analyze how diverse human mobility patterns influence electricity consumption, aiming to provide insights into the complex energy-behavioral relationships.
    
\end{itemize}

\subsection{Related Work} \label{subsec:intro_related_work}

\paragraph{Statistical Methods}
Statistical methods, such as ARIMA, have historically been foundational in load forecasting. However, conventional models often struggle with non-linear data patterns due to their linearity assumptions, limiting accuracy in scenarios with complex load dynamics~\mbox{\cite{Kuster2017,Waheed2024}}. Consequently, recent research has increasingly shifted toward deep learning alternatives.

\paragraph{Deep Learning-based Methods}
Benefiting from the capacity to model complex non-linearities, various deep learning architectures have been adopted. RNNs, particularly LSTM~\mbox{\cite{lstm}} and GRU~\mbox{\cite{gru}}, are widely used to capture temporal dependencies, often utilizing self-feedback mechanisms~\mbox{\cite{Xia2021}}. CNNs, such as TCN~\mbox{\cite{tcn}}, offer effective feature extraction from the temporal dimension via dilated causal convolutions~\mbox{\cite{Wang2024}}. To address spatial dependencies, GNNs have been employed to extract embeddings from power grid topologies~\mbox{\cite{Mansoor2024,Giamarelos2024}} or capture spatial-temporal relationships at multiple scales using graph attention~\mbox{\cite{li2025multi}}. More recently, transformer architectures have become prominent for capturing long-range dependencies; examples include LDTformer, which incorporates hybrid time-frequency attention~\mbox{\cite{Peng2025}}, and encoder-decoder structures designed for diverse temporal patterns~\mbox{\cite{Jiang2024_att}}.

\paragraph{Hybrid Deep Learning Methods}
To leverage the distinct strengths of individual architectures, hybrid models have been extensively developed. One common strategy combines CNNs for feature extraction with RNNs for sequence modeling, such as feeding CNN-extracted features into bidirectional LSTMs~\mbox{\cite{Eskandari2021}} or using Conv2D-GRU structures to identify steep load changes~\mbox{\cite{Chen2024}}. Another approach integrates GNNs with other modules; for instance, SmartFormer integrates GNN components within transformer layers~\mbox{\cite{Saeed2025}}, while other frameworks combine graph convolutions with dendritic neural models~\mbox{\cite{Zhang2025_graph}} or attention mechanisms~\mbox{\cite{Su2024}} to enhance spatial-temporal processing. Furthermore, transformers are frequently hybridized with CNNs~\mbox{\cite{Tian2024,Xu2024}} or LSTMs~\mbox{\cite{Pentsos2025,Dong2025}} to improve feature preprocessing and residual modeling~\mbox{\cite{Hu2024}}.

\paragraph{Interpretable Methods}
Compared to black-box deep-learning methods, interpretable neural forecasting methods, such as attention-based explainable transformer \mbox{\cite{att_interpretable_trans2}} and symbolic regression network \mbox{\cite{symbolic_regression2}}, provide human-understandable insights into the forecasting results. Recently, the KAN~\mbox{\cite{kan}} has emerged as a promising alternative with noticeable advantages. For example, the explainable transformers rely heavily on attention maps which only provide feature importance (i.e., a scalar indicating where to look) but fail to describe the functional relationship, i.e., how the features impact the output, e.g., positive vs. negative correlation, saturation, or quadratic growth, while the symbolic networks suffer from combinatorial complexity and sensitivity to noise. Unlike networks with fixed node activation functions, KAN employs learnable activation functions (namely splines) on edges, offering potential advantages in both function approximation and interpretability. However, the direct application of KAN to load forecasting faces challenges. Recent studies indicate that pure KAN structures often struggle with time-series characteristics, suffering from training instability and performance degradation as depth increases~\mbox{\cite{Han2025,Jiang2025}}. While some initial attempts used pure KANs~\mbox{\cite{Abbas2025}}, results suggest significant architectural modifications are required for effective temporal forecasting.

Consequently, current research focuses on integrating KAN into existing architectures, typically by replacing MLP layers. For example, KAN layers have been inserted into DLinear~\mbox{\cite{Jiang2024}}, TCN, and Transformer frameworks~\mbox{\cite{Zhang2025}} to enhance accuracy. Other approaches adapt KAN for temporal modeling via recurrent structures~\mbox{\cite{Danish2025}} or integrate it with multi-head self-attention~\mbox{\cite{Wang2024_kan}} and CNNs~\mbox{\cite{Pei2025}}. While these hybrids improve forecasting performance, merely using KAN as a replacement module often fails to unlock its full interpretability potential. This creates a specific methodological gap: utilizing KAN not just for accuracy, but to preserve and interpret feature-wise dynamics, a gap that this study aims to fill.

\subsection{Contributions} \label{subsec:intro_contributions}
The above research gaps regarding KAN implementation can be summarized into two aspects: 1) the challenges of pure KAN structures in handling temporal dynamics and 2) the limited and diminishing interpretive depth when KAN is merely used to replace MLP components. To address them, this study develops LoadKAN -- a novel hybrid and interpretable deep learning framework integrated with KAN. It uniquely designs a feature-isolated temporal attention mechanism with a KAN module. The attention stage is engineered to process and extract temporal representations from each input feature sequence independently without entangling its distinct dynamics. These features include historical load, weather, and various human mobility metrics such as residential and workplace presence. The distilled and temporally-aware feature representations are then fed into the KAN module to perform the final load prediction. The architectural design of LoadKAN not only aims for accurate forecasting but is also strategically geared towards comprehensively exploring and interpreting the learned non-linear relationships between the processed input features and the electricity load, with a particular focus on deciphering the hidden interplay between diverse human mobility patterns and energy consumption.

The main contributions of this paper are summarized as follows:
\begin{itemize}
    \item We explore the potential of KAN-based neural models in performing electricity load forecasting and providing interpretable results, through comprehensive evaluation using case studies across representative U.S. electricity markets with significant differences in terms of input features.

    \item We develop the LoadKAN framework, a hybrid deep learning model that synergistically integrates a non-entangled temporal attention mechanism for feature-wise temporal pattern extraction with a KAN module for interpretable prediction. Unlike black-box neural models, LoadKAN enables granular analysis of how individual features affect forecast outcomes.
    Our experimental results demonstrate the effectiveness of our LoadKAN compared to established deep learning benchmark models across all three tested electricity markets and evaluation metrics.
    
    \item We discover interpretable insights by analyzing the impacts of human mobility behaviors on electricity load. First, our ablation study highlights the substantial positive impact of incorporating human mobility features across various models and markets. Second, in addition to KAN-visualized relationships between mobility features and electricity load, we provide an independent, complementary sensitivity analysis to quantify the relative significance of these identified relationships across the three electricity markets. This analysis reveals interesting market-specific dependencies, such as the varying influence of outside activities in public places versus commercial mobility across different jurisdictions of U.S. electricity markets.

\end{itemize}
The remainder of this paper is organized as follows. Section \ref{sec:dataset_task_formulation} details the used electricity load datasets, the ancillary features including human mobility data, and the formal problem definition for load forecasting. Section \ref{sec:loadkan_framework} presents the developed LoadKAN framework, elaborating on its non-entangled temporal feature processing stage and the KAN module. Section \ref{sec:exp} describes the experimental setup, benchmark models, evaluation metrics, and presents a comprehensive analysis of the results, including performance comparisons, the impact of mobility features, and the interpretable analysis based on LoadKAN.

\section{Load Dataset and Forecasting Task Formulation} \label{sec:dataset_task_formulation}
This section first outlines the datasets for electricity load forecasting in Section \ref{subsec:data}, followed by Section \ref{subsec:model_task_formulation}, where the load forecasting task is formally defined.

\subsection{Dataset Overview} \label{subsec:data}
This study uses electricity load data and a range of ancillary features collected from three major U.S. electricity markets: the New York Independent System Operator (NYISO), the California Independent System Operator (CAISO), and the Electric Reliability Council of Texas (ERCOT).

\begin{figure}[!t]
    \centering
    \includegraphics[width=\linewidth]{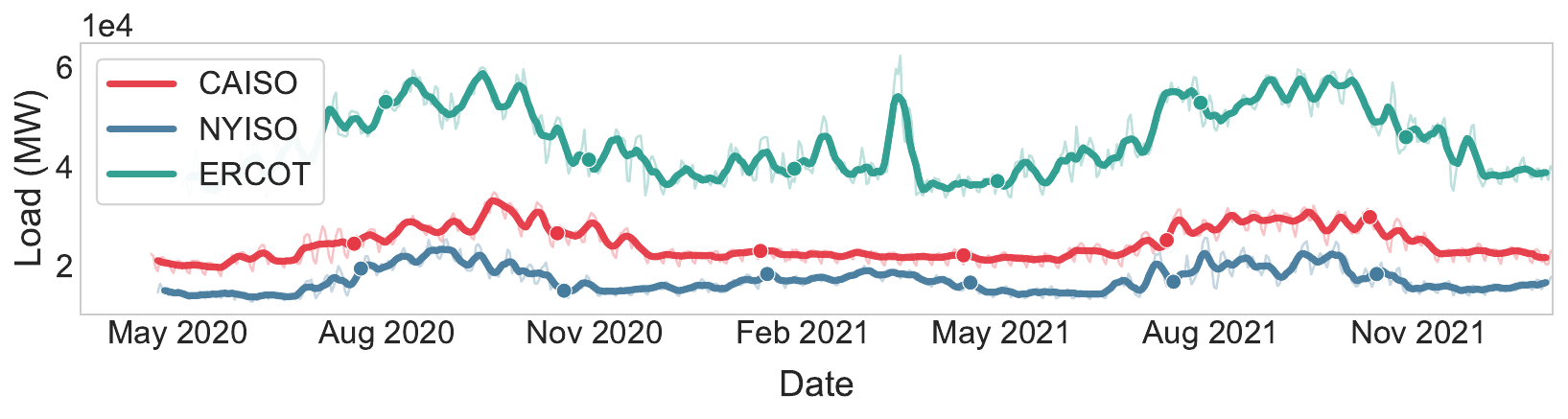}
    \caption{Load profiles of NYISO, CAISO, and ERCOT.}
    \label{fig:load_visualization}
\end{figure}

\paragraph{Electricity Load Data}
The electricity load data is sourced from the COVID-EMDA+ data hub~\cite{ruan_2020}. Fig. \ref{fig:load_visualization} illustrates the load profiles for NYISO, CAISO, and ERCOT from May 2020 to November 2021. As depicted, ERCOT generally exhibits the highest load levels and considerable seasonal variation, followed by CAISO and NYISO, each displaying unique demand characteristics influenced by regional climates and consumer behaviors. These profiles highlight the diverse operational scales and temporal patterns across the markets.

\paragraph{Weather and Market Price}
To improve load forecasting accuracy, the load data is augmented with several categories of ancillary features. These include key meteorological variables from the respective market jurisdictions, such as dewpoint temperature, relative humidity, wind speed, and ambient temperature. Additionally, locational marginal price data for each electricity market is incorporated as an economic indicator that can influence electricity consumption.

\begin{figure}[!t]
    \centering
    \includegraphics[width=\linewidth]{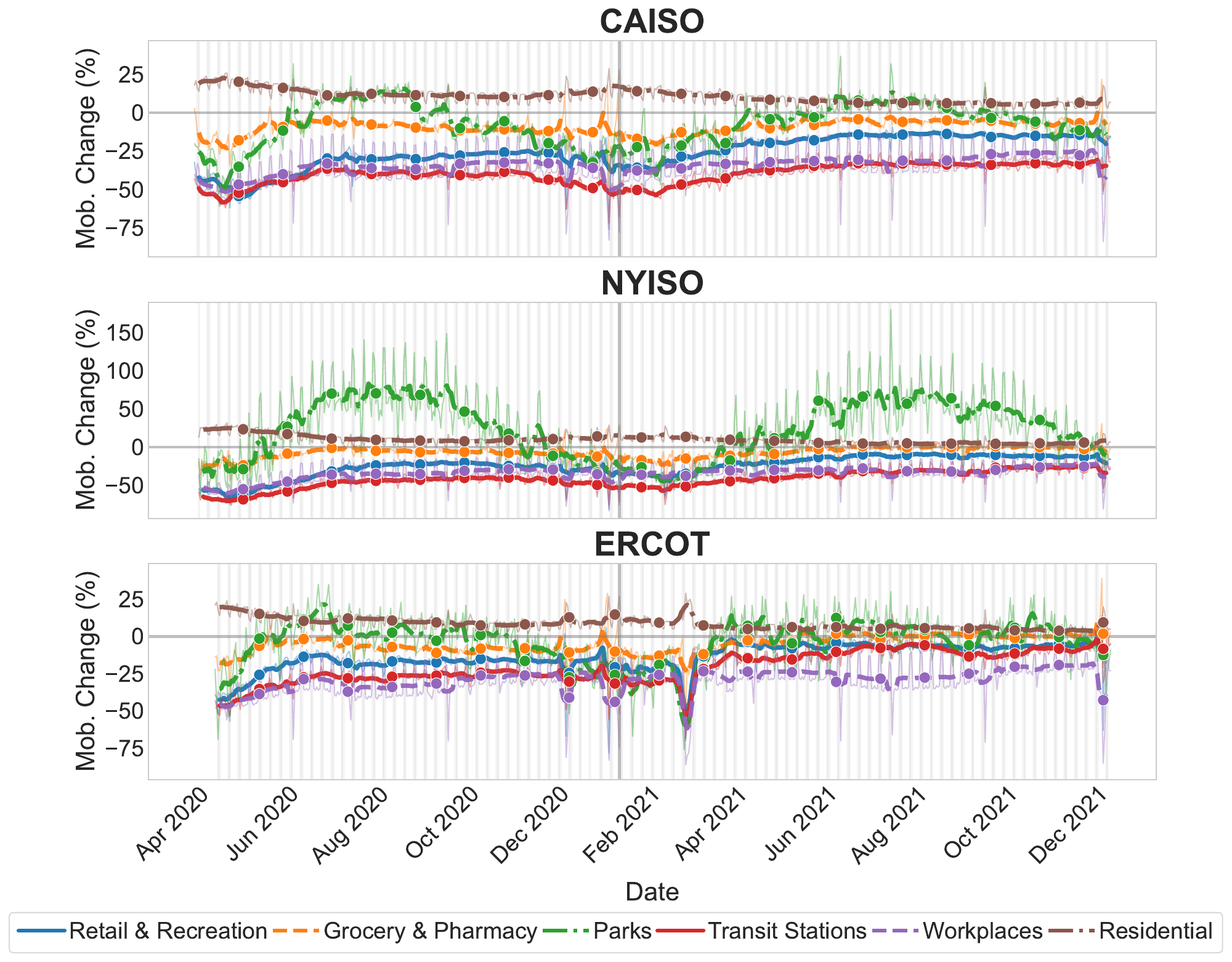}
    \caption{Mobility features of CAISO, NYISO, and ERCOT.}
    \label{fig:vis_feat_mobility}
\end{figure}

\paragraph{Human Mobility Data} Recognizing the significant impact of human presence and activity on electricity demand~\cite{Chen2020,Liu2022}, this study includes human mobility data sourced from Google's COVID-19 Community Mobility Reports (CCMR)~\cite{google_2021}. The CCMR dataset quantifies daily mobility trends across six distinct categories:
\begin{enumerate}
    \item \textbf{Retail \& Recreation:} Mobility trends for places like restaurants, cafes, libraries, and movie theaters.
    \item \textbf{Grocery \& Pharmacy:} Mobility trends for places like grocery markets, food warehouses, farmers markets, specialty food shops, drug stores, and pharmacies.
    \item \textbf{Parks:} Mobility trends for places like national parks, public beaches, marinas, dog parks, plazas, and public gardens.
    \item \textbf{Transit Stations:} Mobility trends for public transport hubs such as subway, bus, and train stations.
    \item \textbf{Workplaces:} Mobility trends for places of work.
    \item \textbf{Residential:} Mobility trends for places of residence.
\end{enumerate}
These mobility metrics are measured as a daily percentage change in visitors (for categories 1-5) or time spent (for category 6) relative to a baseline period -- the median value from the 5-week period from January 3rd to February 6th, 2020. Fig. \ref{fig:vis_feat_mobility} provides a visualization of these six mobility features for CAISO, NYISO, and ERCOT from approximately April 2020 to December 2021, clearly showing distinct regional variations.

Note that all features described above are aggregated to a daily resolution. This daily aggregated dataset forms the foundation for the forecasting task formulated next.

\subsection{Task Formulation} \label{subsec:model_task_formulation}
The objective of this study is to develop a model for accurate multi-step daily electricity load forecasting, leveraging the daily aggregated features outlined in Section \ref{subsec:data}.
Given a historical sequence of $M$ input features up to the current time step $t$, denoted as $\mathbf{x}_t = [\mathbf{f}_{t-L+1}, \dots, \mathbf{f}_t]$, where each $\mathbf{f}_{\tau} \in \mathbb{R}^{M}$ represents the vector of $M$ input features at daily time step $\tau$, the task is to predict the future electricity load sequence $\hat{\mathbf{y}}_t = [\hat{y}_{t+1}, \dots, \hat{y}_{t+H}] \in \mathbb{R}^{H}$. Here, $L$ represents the lookback window size (in days) and $H$ is the forecast horizon (in days). Note that, to ensure strict forecasting realism and prevent data leakage, all exogenous inputs, including weather, human mobility, and locational marginal prices, are strictly lagged. At inference time (e.g., time step $t$), the model strictly relies on historical observations up to $t$ to forecast the multi-step horizon from $t+1$ to $t+H$. No future values of these exogenous features are assumed to be available.

The input data for training the model consists of a set of historical sequences and their corresponding future load values. Let the training dataset be denoted as $\mathcal{D}_{\text{train}} = \{(\mathbf{x}^{(i)}, \mathbf{y}^{(i)})\}_{i=1}^{N_{\text{train}}}$, where $\mathbf{x}^{(i)} = [\mathbf{f}^{(i)}_{t_i-L+1}, \dots, \mathbf{f}^{(i)}_{t_i}]$ is the $i$-th training input sequence and $\mathbf{y}^{(i)} = [y^{(i)}_{t_i+1}, \dots, y^{(i)}_{t_i+H}]$ is the corresponding $H$-step future load sequence. The test dataset is denoted similarly as $\mathcal{D}_{\text{test}} = \{(\mathbf{x}^{(i)}, \mathbf{y}^{(i)})\}_{i=1}^{N_{\text{test}}}$.

The LoadKAN model, denoted by $\mathcal{M}(\mathbf{x}_t; \Theta)$, aims to learn a mapping from the input sequence $\mathbf{x}_t$ to the predicted future load sequence $\hat{\mathbf{y}}_t$, where $\Theta$ represents the set of all learnable parameters within the model. The output of the model for a given input $\mathbf{x}_t$ is written as  $\hat{\mathbf{y}}_t = \mathcal{M}(\mathbf{x}_t; \Theta) = [\hat{y}_{t+1}, \dots, \hat{y}_{t+H}]$.

The model's parameters $\Theta$ are learned by minimizing the mean squared error (MSE) between the predicted load sequence and the actual future load sequence over the training dataset $\mathcal{D}_{\text{train}}$. The MSE loss function for a single training sample $(\mathbf{x}^{(i)}, \mathbf{y}^{(i)})$ is defined as
\begin{equation}
    \mathcal{L}(\mathbf{y}^{(i)}, \hat{\mathbf{y}}^{(i)}) = \frac{1}{H} \sum_{h=1}^{H} (y^{(i)}_{t_i+h} - \hat{y}^{(i)}_{t_i+h})^2.
\end{equation}
The overall objective is to find the optimal set of parameters $\Theta^*$ that minimizes the average MSE loss over the entire training dataset, expressed as
\begin{equation}
    \Theta^* = \underset{\Theta}{\operatorname{argmin}} \frac{1}{N_{\text{train}}} \sum_{i=1}^{N_{\text{train}}} \mathcal{L}(\mathbf{y}^{(i)}, \mathcal{M}(\mathbf{x}^{(i)}; \Theta)).
\end{equation}

While sub-hourly forecasting is crucial for real-time dispatch, the daily-resolution forecasting developed in this study serves essential, macro-level operational and planning use cases. Accurate multi-day daily forecasts are essential for independent system operators (ISOs) and generation companies to perform unit commitment of slow-starting baseload generators, schedule sufficient operating reserves and long-duration energy storage operations, and plan critical system maintenance. Furthermore, since the human mobility metrics utilized in this study are inherently aggregated daily, they act as robust indicators of macro-level behavioral shifts.

\section{Load Forecast Framework of LoadKAN} \label{sec:loadkan_framework}

\begin{figure}
    \centering
    \includegraphics[width=\linewidth]{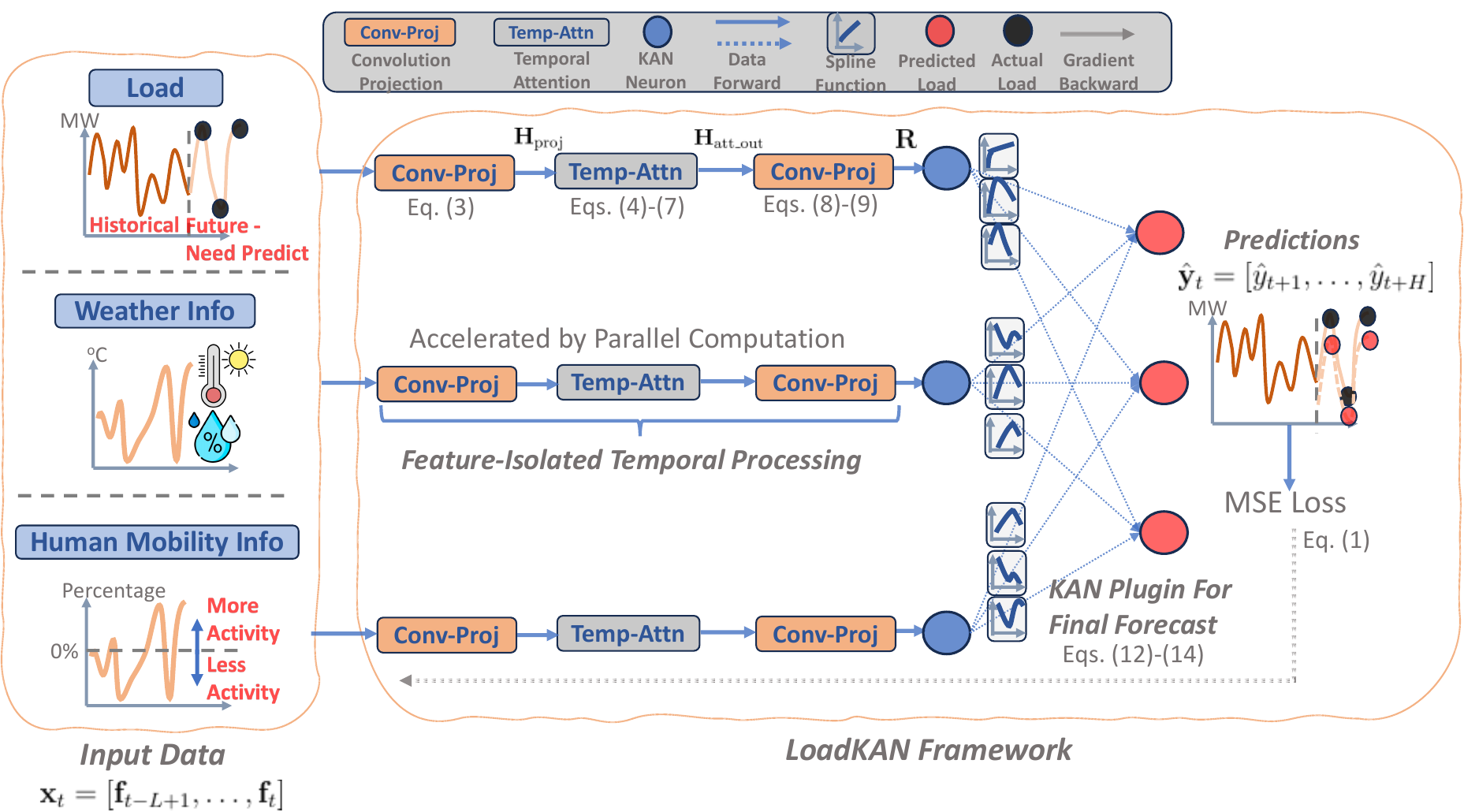}
    \caption{The LoadKAN framework.}
    \label{fig:loadkan_architecture}
\end{figure}

The LoadKAN framework is engineered for accurate electricity load forecasting, combining temporal feature extraction with interpretability and expressive capacity via KAN. As illustrated in Fig. \ref{fig:loadkan_architecture}, the architecture proceeds in two principal stages: first, a non-entangled temporal feature processing stage captures the dynamics of each input feature sequence independently; second, a KAN module models the relationships between these processed feature representations and the final load forecast. The following section provides a rigorous mathematical formulation of each module within LoadKAN.

For a specific forecast initiation time $t$, the input to the LoadKAN model can be expressed as $\mathbf{x}_t = [\mathbf{f}_{t-L+1}, \dots, \mathbf{f}_t]$, where each $\mathbf{f}_{\tau} \in \mathbb{R}^{M}$ represents the vector of $M$ input features (i.e., historical load and exogenous variables like mobility metrics) at time step $\tau$. Thus, $\mathbf{x}_t$ can be viewed as a matrix in $\mathbb{R}^{L \times M}$. For batch processing, the input can be written as $\mathbf{X}_{\text{batch}} \in \mathbb{R}^{B \times L \times M}$, where $B$ is defined as the batch size. The LoadKAN's objective is to predict the future load sequence $\hat{\mathbf{y}}_t = [\hat{y}_{t+1}, \dots, \hat{y}_{t+H}] \in \mathbb{R}^{H}$.

\subsection{Non-Entangled Temporal Feature Processing via Attention Mechanism} \label{subsec:temporal_processing}

This initial stage aims to independently process each of the $M$ input features over the lookback window $L$. In this study, ``non-entangled'' refers to the design in which each input feature sequence is processed independently in the temporal attention stage before any cross-feature interaction is introduced. The goal is to derive a condensed and temporally-aware representation for each feature before these representations are collectively interpreted by the subsequent KAN module. This feature-isolated approach ensures that the unique temporal dynamics of each input stream are distinctly captured.

\paragraph{Feature-Isolated Input Projection}
The batch input $\mathbf{X}_{\text{batch}} \in \mathbb{R}^{B \times L \times M}$ is first transposed to $\mathbf{X}'_{\text{batch}} \in \mathbb{R}^{B \times M \times L}$, ensuring the alignment with the channel-first convention for 1D convolutions. Each of the $M$ feature sequences is then independently projected into a higher-dimensional embedding space of dimension $D_h$, representing hidden dimension per feature. This projection is performed by a 1D convolutional layer configured with $M$ input channels, $M \cdot D_h$ output channels, a kernel size of 1, and $M$ groups.
For a single instance $\mathbf{x}'_t \in \mathbb{R}^{M \times L}$ from $\mathbf{X}'_{\text{batch}}$, let $(\mathbf{x}'_t)_{m,\ell}$ be the value of the $m$-th feature at time step $\ell$ within the lookback window ($\ell \in \{1, \dots, L\}$). The projection for the $m$-th feature group and its $d$-th output channel ($d \in \{1, \dots, D_h\}$) at time step $\ell$ can be formulated as
\begin{equation}
\label{eq:feature_isolated_input_proj}
    (\mathbf{H}_{\text{proj}})_{b, (m-1)D_h+d, \ell} = W^{\text{in}}_{m,d} \cdot (\mathbf{X}'_{\text{batch}})_{b,m,\ell} + B^{\text{in}}_{m,d},
\end{equation}
where $W^{\text{in}}_{m,d}$ and $B^{\text{in}}_{m,d}$ are the scalar learnable weight and bias for the $d$-th filter of the $m$-th feature group. The output of this layer is denoted as $\mathbf{H}_{\text{proj}} \in \mathbb{R}^{B \times (M \cdot D_h) \times L}$. Eq. \mbox{\eqref{eq:feature_isolated_input_proj}} expands the neural model's capacity, enabling it to encode localized context for each feature independently before temporal processing begins.

\paragraph{Reshaping and Positional Encoding}
The projected sequences $\mathbf{H}_{\text{proj}}$ are reshaped and permuted to $\mathbf{H}_{\text{att\_in}} \in \mathbb{R}^{(B \cdot M) \times L \times D_h}$. Such transformation effectively creates a batch where each of the $M$ features from each of the $B$ samples is treated as an independent sequence of length $L$ and dimension $D_h$ for the attention mechanism.
Standard sinusoidal positional encoding \cite{transformer} $\mathbf{P} \in \mathbb{R}^{L \times D_h}$ is added element-wise to these sequences, formulated as
\begin{equation}
\label{eq:pos_enc_1}
    \mathbf{H}'_{\text{att\_in}} = \mathbf{H}_{\text{att\_in}} + \mathbf{P},
\end{equation}
In Eq. \eqref{eq:pos_enc_1}, the components of $\mathbf{P}$ are defined for position $l \in \{0, \dots, L-1\}$ and dimension index $k$ as
\begin{equation}
\label{eq:pos_enc_2}
    P_{l, 2k} = \sin(l / \theta^{2k/D_h}), \quad P_{l, 2k+1} = \cos(l / \theta^{2k/D_h}),
\end{equation}
where $\theta$ is a large constant set to a default value of $10000$. Eq. \mbox{\eqref{eq:pos_enc_1}} and \mbox{\eqref{eq:pos_enc_2}} ensure the model considers the chronological sequence of the input data, allowing it to distinguish between recent trends and distant historical patterns.

\paragraph{Feature-Wise Multi-Head Self-Attention}
The positionally encoded sequences $\mathbf{H}'_{\text{att\_in}}$ are processed by a stack of $N_{\text{att\_layers}}$ transformer encoder layers. Each layer comprises a multi-head self-attention (MHSA) mechanism and a position-wise feed-forward network (FFN), with layer normalization and residual connections.
For each sequence $\mathbf{z} \in \mathbb{R}^{L \times D_h}$ in $\mathbf{H}'_{\text{att\_in}}$, the MHSA output is written as
\begin{equation}
\label{eq:MHSA_overall}
    \text{MHSA}(\mathbf{z}) = \text{Concat}(\text{head}_1, \dots, \text{head}_{N_h}) \mathbf{W}^O.
\end{equation}
Each head$_j$ is calculated as
\begin{equation}
\label{eq:MHSA_head}
    \begin{aligned}
        \text{head}_j &= \text{Attention}(\mathbf{z}\mathbf{W}^Q_j, \mathbf{z}\mathbf{W}^K_j, \mathbf{z}\mathbf{W}^V_j), \\ 
        & = \text{softmax}\left(\frac{(\mathbf{z}\mathbf{W}^Q_j)(\mathbf{z}\mathbf{W}^K_j)^T}{\sqrt{d_k}}\right) (\mathbf{z}\mathbf{W}^V_j),
    \end{aligned}
\end{equation}
where $\mathbf{W}^Q_j, \mathbf{W}^K_j, \mathbf{W}^V_j$ are learnable projection matrices for queries, keys, and values for head $j$, $N_h$ is the number of heads, $d_k$ is the dimension of keys per head, and $\mathbf{W}^O$ is an output projection matrix. The FFN is set as a two-layer MLP. This processing allows each feature's temporal sequence to attend to its own past values, capturing intricate temporal dependencies independently. The output of the MHSA is denoted as $\mathbf{H}_{\text{att\_out}} \in \mathbb{R}^{(B \cdot M) \times L \times D_h}$. As shown in Eqs. \mbox{\eqref{eq:MHSA_overall}} and \mbox{\eqref{eq:MHSA_head}}, our MHSA, unlike standard architectures that entangle all features, processes each feature's history independently, which acts as a temporal filter, identifying and weighting the most critical historical moments for that specific feature while suppressing irrelevant noise.

\paragraph{Temporal Information Aggregation and Output Projection}
The temporally processed information for each feature is aggregated by selecting the output corresponding to the last time step from $\mathbf{H}_{\text{att\_out}}$, expressed as $\mathbf{h}_{\text{last}} = \mathbf{H}_{\text{att\_out}}[:, -1, :] \in \mathbb{R}^{(B \cdot M) \times D_h}$. This choice is consistent with the forecasting formulation, where predictions are initiated from the most recently observed time step $t$. Since the final output of the self-attention module attends to the entire lookback window, it provides a compact feature-specific representation that summarizes informative historical signals conditioned on the latest state. This output $\mathbf{h}_{\text{last}}$ is further reshaped to $\mathbf{H}_{\text{featwise}} \in \mathbb{R}^{B \times M \times D_h}$, which is further converted into $\mathbf{H}_{\text{conv\_in}} \in \mathbb{R}^{B \times (M \cdot D_h) \times 1}$ to be processed by a final feature-isolated 1D convolution. This layer has $M \cdot D_h$ input channels, $M$ output channels, kernel size 1, and $M$ groups, applying an independent linear transformation to each feature's $D_h$-dimensional aggregated temporal representation. After such an aggregation layer, a single scalar value $r'_{m}$ for each feature $m$ is finally obtained and expressed as
\begin{equation}
\label{eq:temporal_aggregation}
    r'_{m} = (\mathbf{w}^{\text{out}}_{m})^T \mathbf{h}_{\text{featwise},m} + b^{\text{out}}_{m},
\end{equation}
where $\mathbf{h}_{\text{featwise},m} \in \mathbb{R}^{D_h}$ represents the aggregated representation for feature $m$, and $\mathbf{w}^{\text{out}}_{m} \in \mathbb{R}^{D_h}, b^{\text{out}}_{m} \in \mathbb{R}$ denote learnable parameters. The collection of these values forms $\mathbf{R}'_{\text{raw}} \in \mathbb{R}^{B \times M}$. Eq. \mbox{\eqref{eq:temporal_aggregation}} condenses the complex historical dynamics of feature $m$ into a single scalar value, effectively summarizing the current state of that feature based on its past trajectory.

\paragraph{Final Processed Feature Representation}
A hyperbolic tangent activation function, defined as $\tanh(\cdot)$, is applied element-wise to obtain the final processed feature representation vector $\mathbf{R} \in \mathbb{R}^{B \times M}$:
\begin{equation}
\label{eq:tanh_restrict}
    \mathbf{R} = \tanh(\mathbf{R}'_{\text{raw}}).
\end{equation}
Each row $\mathbf{r}^{(b)} = [r_1^{(b)}, r_2^{(b)}, \dots, r_M^{(b)}]$ of $\mathbf{R}$ represents the input to the KAN layer, where $r_m^{(b)}$ is the distilled and temporally-aware representation of the $m$-th original input feature for batch sample $b$. Eq. \mbox{\eqref{eq:tanh_restrict}} normalizes the aggregated features into the range of $[-1,1]$. Since the following KAN layer operates over a bounded domain, this normalization is crucial for handling real-world data outliers and ensuring the interpretable layer receives consistent, stable inputs.

The design choice to process features in isolation before aggregation is driven by the need to prevent both signal dominance and temporal blurring. Within this design, the feature-wise attention weights are learned to identify informative historical time steps within each individual feature sequence, thereby allowing the model to emphasize the most forecast-relevant temporal moments of each feature while suppressing less informative variations. By employing a feature-isolated attention mechanism, we effectively decouple the temporal processing of each variable. Such a tailored mechanism allows our framework to act as a collection of specialized experts: one attention head focuses exclusively on extracting the periodicity of the load curve, while another parallel head independently identifies the unique lag structures of human mobility patterns. This architectural constraint ensures that the unique temporal signature of every input feature is preserved and refined independently, preventing feature cross-talk until the final interpretable KAN layer.  

More importantly, this isolation is an essential prerequisite for the subsequent interpretable KAN analysis. In standard attention architectures, the mixing of features creates entangled latent embeddings. If a KAN layer were applied to such embeddings, the learned functions would describe relationships with abstract latent variables, not the physical drivers. By strictly prohibiting feature interaction until the final aggregation stage, our mechanism successfully prevents this entanglement. This ensures that the input to each KAN spline remains a pure representation of the specific real-world feature (e.g., mobility), guaranteeing that the visualized functions reflect true physical dependencies rather than opaque model artifacts.

\subsection{KAN Module for Final Forecast} \label{subsec:kan_plugin}

As illustrated in Fig. \ref{fig:loadkan_architecture}, the KAN module acts as the final forecasting module, mapping the processed feature representations $\mathbf{R} \in \mathbb{R}^{B \times M}$ from the attention stage to the $H$-step ahead load forecast $\hat{\mathbf{Y}} \in \mathbb{R}^{B \times H}$. This KAN layer is constructed using learnable B-spline activation functions, enabling it to model non-linear interactions between the processed features and the output predictions in an inherently interpretable manner.

\paragraph{B-Spline Basis Functions}
The foundation of KAN's adaptive activations lies in a set of B-spline basis functions~\cite{kan}. For an input scalar $x$ (an element $r_m^{(b)}$ from $\mathbf{R}$) within a defined domain $[d_{\min}, d_{\max}]$, B-splines of degree $p$ are constructed over a knot vector $\mathbf{t} = (t_0, t_1, \dots, t_{N_b+p})$, where $N_b$ is the number of basis functions. The $k$-th B-spline basis function, $B_{k,p}(x)$, is defined by the Cox-de Boor recursion formula and formulated as
\begin{equation}
\label{eq:bspline_init}
     B_{k,0}(x) = \begin{cases} 1 & \text{if } t_k \le x < t_{k+1}, \\ 0 & \text{otherwise} \end{cases}
\end{equation}
For $p > 0$, we have the following recursive expression
\begin{equation}
\label{eq:bspline_general}
    B_{k,p}(x) = \frac{x - t_k}{t_{k+p} - t_k} B_{k,p-1}(x) + \frac{t_{k+p+1} - x}{t_{k+p+1} - t_{k+1}} B_{k+1,p-1}(x).
\end{equation}
We handle coincident knots by setting terms to zero. The B-spline module computes a vector (with the length of $N_b$) of basis function values -- $\bm{\phi}(x) = [B_{0,p}(x), \dots, B_{N_b-1,p}(x)]^T$. When applied to each element $r_m^{(b)}$ of the input representation $\mathbf{R}$, it yields basis values $\mathbf{\Phi} \in \mathbb{R}^{B \times M \times N_b}$, where $\Phi_{b,m,k'} = B_{k'-1,p}(r_m^{(b)})$ for $k' \in \{1, \dots, N_b\}$. Unlike standard neural networks that use fixed linear weights, Eqs. \mbox{\eqref{eq:bspline_init}} and \mbox{\eqref{eq:bspline_general}} enable our neural model to construct flexible, learnable curves, thereby capturing highly non-linear relationships between the input features and the electricity load, such as threshold effects or saturation points.

\paragraph{KAN Layer Formulation}
The KAN layer takes the processed feature representations $\mathbf{R} \in \mathbb{R}^{B \times M}$ as input and produces the forecast $\hat{\mathbf{Y}} \in \mathbb{R}^{B \times H}$. For each output forecast step $h \in \{1, \dots, H\}$, the prediction $\hat{y}_{t+h}^{(b)}$ for batch sample $b$ (corresponding to input $\mathbf{x}_t^{(b)}$) is modeled as:
\begin{equation}
\label{eq:kan_layer_formulation}
    \hat{y}_{t+h}^{(b)} = \left( \sum_{m=1}^{M} f_{m,h}(r_m^{(b)}) \right) + \beta_h,
\end{equation}
where $r_m^{(b)}$ is the $m$-th processed input feature for sample $b$, $\beta_h$ is a learnable bias term for the $h$-th output step, and each univariate function $f_{m,h}(\cdot)$ is approximated by a linear combination of B-spline basis functions, expressed as
\begin{equation}
\label{eq:kan_layer_inside}
    f_{m,h}(r_m^{(b)}) = \sum_{k'=1}^{N_b} w_{m,h,k'} \cdot B_{k'-1,p}(r_m^{(b)}),
\end{equation}
where $w_{m,h,k'}$ are the learnable spline coefficients. Each function $f_{m,h}(r_m^{(b)})$ represents a learnable activation function specific to the $m$-th processed feature when predicting the $h$-th future load value. Hence, the KAN layer operation for the $h$-th forecast step can be further broken down into
\begin{equation}
\label{eq:kan_layer_final_formulation}
    \hat{y}_{t+h}^{(b)} = \left( \sum_{m=1}^{M} \sum_{k'=1}^{N_b} w_{m,h,k'} \cdot \Phi_{b,m,k'} \right) + \beta_h.
\end{equation}
Eq. \mbox{\eqref{eq:kan_layer_final_formulation}} replaces the opaque matrix multiplications of ``black-box'' neural models with a transparent, additive structure, enabling the visualization and interpretation of exactly how a change in a single input feature impacts the final load forecast.

In summary, the hybrid and feature-isolated structure of LoadKAN enables learning distinct and highly non-linear relationships for each processed input feature, as well as explaining how it contributes to each step's prediction in the forecast horizon. The feature-isolated temporal processing stage is essential to ensuring that these final relationships are learned based on individually refined feature representations without entangled dynamics. The algorithmic procedure of our LoadKAN is provided in Algorithm \mbox{\ref{alg:loadkan}}.

\begin{remark}[KAN Interpretability and Expressivity]
    
    The interpretability of the KAN plugin comes from the fundamental additive decomposition form of the Kolmogorov-Arnold Representation Theorem. The theorem posits that any multivariate continuous function defined on a bounded domain can be represented as a finite sum of continuous univariate functions, creating the additivity attribute for KAN. While the Kolmogorov-Arnold Representation Theorem requires a specific nested structure to achieve universality for any continuous function, the original work of KAN \cite{kan} demonstrates that even shallow or constrained KAN structures, like Eq. \eqref{eq:kan_layer_formulation} in our study, are still powerful approximators.
    
    Additionally, the additive nature of KAN does not compromise its expressivity. By mapping each processed feature into multiple learnable B-spline basis functions with dynamic, learnable coefficients, the model maintains high expressive capacity in capturing highly non-linear, smooth patterns like threshold effects or saturation points that simpler additive models, such as ReLU-based sub-networks relying on piecewise-linear approximations \cite{noorizadegan2025practitionersguidekolmogorovarnoldnetworks,zeng2024kanversusmlpirregular,cang2024kanworkexploringpotential}, might miss.

\end{remark}

\begin{algorithm}[!t]
\caption{Training Process of the LoadKAN Framework}
\label{alg:loadkan}
\begin{algorithmic}[1]
\Require Training dataset $\mathcal{D}_\mathrm{train} = \{(x^{(i)}, y^{(i)})\}_{i=1}^{N_\mathrm{train}}$, Lookback window $L$, Forecast horizon $H$, Number of features $M$, Hidden dimension $D_h$, Number of B-spline basis functions $N_b$, Degree of splines $p$, Batch size $B$.
\Ensure Optimal model parameters $\Theta^*$.
\State Initialize model parameters $\Theta$ randomly.
\While{not converged or max epochs not reached}
    \State Sample a mini-batch of inputs $X_\mathrm{batch} \in \mathbb{R}^{B \times L \times M}$ and targets $Y_\mathrm{batch} \in \mathbb{R}^{B \times H}$.
    
    \Comment{\textbf{Stage 1: Feature-Isolated Temporal Attention}}
    \State Transpose input to $X'_\mathrm{batch} \in \mathbb{R}^{B \times M \times L}$.
    \State Apply feature-isolated 1D convolution to obtain $H_\mathrm{proj} \in \mathbb{R}^{B \times (M \cdot D_h) \times L}$.
    \State Reshape $H_\mathrm{proj}$ and add positional encoding $P$ to obtain $H'_\mathrm{att\_in} \in \mathbb{R}^{(B \cdot M) \times L \times D_h}$.
    \State Apply feature-wise Multi-Head Self-Attention (MHSA) to get $H_\mathrm{att\_out}$.
    \State Extract the last time step for temporal aggregation: $h_\mathrm{last} = H_\mathrm{att\_out}[:, -1, :]$.
    \State Apply output 1D convolution to $h_\mathrm{last}$ to yield $R'_\mathrm{raw} \in \mathbb{R}^{B \times M}$.
    \State Apply activation to get final processed representations: $R = \tanh(R'_\mathrm{raw})$.
    
    \Comment{\textbf{Stage 2: KAN Module for Final Forecast}}
    \State Compute B-spline basis function values $\Phi_{b,m,k'} = B_{k'-1,p}(r^{(b)}_m)$ for all items in $R$.
    \State Compute $H$-step ahead predictions: $\hat{y}_{t+h}^{(b)} = (\sum_{m=1}^{M} \sum_{k'=1}^{N_b} w_{m,h,k'} \cdot \Phi_{b,m,k'}) + \beta_h$.
    
    \Comment{\textbf{Optimization}}
    \State Compute the MSE loss: $\mathcal{L} = \frac{1}{B \cdot H}\sum_{b=1}^{B}\sum_{h=1}^{H}(y_{t+h}^{(b)} - \hat{y}_{t+h}^{(b)})^2$.
    \State Update model parameters $\Theta$ using the Adam optimizer to minimize $\mathcal{L}$.
\EndWhile
\State \Return $\Theta^*$
\end{algorithmic}
\end{algorithm}

\section{Experiments} \label{sec:exp}

\subsection{Experimental Settings} \label{subsec:exp_settings}

This subsection outlines the framework for our empirical evaluation, detailing the benchmark models used for comparison, the metrics for assessing performance, and the comprehensive setup of our experiments.

\subsubsection{Benchmarks} \label{subsubsec:exp_settings_benchmarks}
To rigorously evaluate the performance of our developed LoadKAN model, we introduce a comparative baseline using six benchmark models, including a direct application of KAN and five widely recognized neural network architectures commonly employed for time-series forecasting tasks, including MLP, LSTM \mbox{\cite{lstm}}, GRU \mbox{\cite{gru}}, TCN \mbox{\cite{tcn}}, Transformer \mbox{\cite{transformer}}, Informer \mbox{\cite{zhou2021informer}}, Chronos \mbox{\cite{ansari2024chronos}}, and PureKAN \mbox{\cite{Abbas2025,Jiang2025}}. Detailed descriptions of these models, as well as hyperparameter configurations are provided in \mbox{\ref{appendix:benchmark_description_and_config}}.

\subsubsection{Evaluation Metrics} \label{subsubsec:exp_settings_metrics} 
To quantitatively assess and compare the performance of the developed LoadKAN model against the benchmarks, we evaluate their predictions on the test set, previously defined as $\mathcal{D}_{\text{test}}$. The evaluation considers the forecasting accuracy over the defined forecast horizon $H$ using three standard metrics: root mean squared error (RMSE), mean absolute percentage error (MAPE), and the coefficient of determination ($\mathrm{R}^2$). The detailed formulations of these metrics are provided in \mbox{\ref{appendix:evaluation_metrics}}.

\subsubsection{Experimental Setup and Configurations} \label{subsubsec:exp_settings_configuration}
This subsection details the experimental setup, including data preparation, training procedures, implementation settings, and the specific configurations for the benchmark models and our developed LoadKAN.

\paragraph{Data Preparation}
The original tabular dataset is divided chronologically into training and test sets, maintaining the temporal order crucial for time series forecasting. The train set comprises the initial $80\%$ of the data, while the remaining $20\%$ constitutes the test set used for final performance evaluation. Each feature within the train and test sets is separately normalized using min-max scaling into the range of $[-1, 1]$. To finally derive the train and test sets -- $\mathcal{D}_\textrm{train}$ and $\mathcal{D}_\textrm{test}$ defined in Section \ref{subsec:model_task_formulation}, we create time-series sequences in a rolling-horizon manner based on the two parameters -- historical input length $L$ and forecast horizon $H$ with default values of $7$ and $3$ in our study, respectively. The 7-day lookback window is selected to capture weekly temporal regularities in daily electricity load, such as weekday-weekend demand differences. The 3-day forecast horizon represents a practical multi-day-ahead forecasting task that is relevant to short-term operational planning.

\paragraph{Training Procedure}
All experiments are performed on an NVIDIA V100 GPU. The implementations are developed using Python (version 3.10.13) and the PyTorch deep learning framework (version 1.12.0). All neural network models, including the benchmarks and LoadKAN, are trained using a consistent procedure to ensure fair comparison. The Adam optimizer is utilized for updating model parameters. The learning rate is set to $0.001$. Models are trained using a batch size of $32$. Training is conducted for a maximum of $300$ epochs. An early stopping mechanism is implemented to mitigate overfitting; training is terminated if the training loss does not show significant improvement for $50$ consecutive epochs. 

\paragraph{Neural Architecture Design}
The architectural specifics for each model, i.e., both baselines and our LoadKAN, are outlined in \mbox{\ref{appendix:benchmark_description_and_config}}.

\subsection{Experimental Results} \label{subsec:exp_results}

This subsection presents the outcomes of our experiments, beginning with an overall forecast performance comparison of LoadKAN against the benchmark models in Section \ref{subsubsec:exp_results_performance}, followed by Section \ref{subsubsec:exp_results_impact_mobility}, where  an ablation study is conducted to quantify the impact of human mobility features. Section \ref{subsubsec:exp_results_interpretable} provides a detailed interpretable analysis of LoadKAN's learned relationships and a discussion of their broader implications.

\begin{table*}[!t]
    \centering
    
    \caption{Load forecast performance metrics (MAPE, RMSE, R$^2$) in NYISO, CAISO, and ERCOT markets with the inclusion of mobility features. The Transformer model is referred to as Trans. for brevity.}
    
    \resizebox{2\columnwidth}{!}{
    \begin{tabular}{ l c c c c c c c c c c}
        \hline
        \textbf{Market} & \textbf{Metric} & \textbf{MLP} & \textbf{LSTM} & \textbf{GRU} & \textbf{TCN} & \textbf{Trans.} & \textbf{Informer} & \textbf{Chronos} & \textbf{PureKAN} & \textbf{LoadKAN} \\
        \hline
        \multirow{3}{*}{NYISO} & MAPE & 4.88\% & 4.24\% & 4.31\% & 3.44\% & 3.51\% & $\bm{3.17\%}$ \textbf{(1st)} & 4.07\% & 9.87\% & $\bm{3.34\%}$ \textbf{(2nd)} \\
         & RMSE (MW) & 1172 & 1047 & 1062 & 861 & 877 & $\bm{812}$ \textbf{(1st)} & 1009 & 2244 & $\bm{847}$ \textbf{(2nd)} \\
         & $R^2$ & 0.776 & 0.814 & 0.807 & 0.908 & 0.903 & $\bm{0.924}$ \textbf{(1st)} & 0.832 & 0.281 & $\bm{0.914}$ \textbf{(2nd)} \\
        \hline \hline
        \multirow{3}{*}{CAISO} & MAPE & 3.72\% & 3.14\% & 3.22\% & $\bm{1.81\%}$ \textbf{(1st)} & 2.09\% & 2.03\% & 2.72\% & 6.47\% & $\bm{1.94\%}$ \textbf{(2nd)} \\
         & RMSE (MW) & 1276 & 1077 & 1098 & $\bm{596}$ \textbf{(1st)} & 682 & 668 & 937 & 1943 & $\bm{641}$ \textbf{(2nd)} \\
         & $R^2$ & 0.864 & 0.903 & 0.897 & $\bm{0.978}$ \textbf{(1st)} & 0.963 & 0.967 & 0.926 & 0.651 & $\bm{0.971}$ \textbf{(2nd)} \\
        \hline \hline 
        \multirow{3}{*}{ERCOT} & MAPE & 4.52\% & 4.11\% & 4.19\% & 3.16\% & $\bm{2.97\%}$ \textbf{(2nd)} & 3.03\% & 3.83\% & 7.82\% & $\bm{2.91\%}$ \textbf{(1st)} \\
         & RMSE (MW) & 2743 & 2446 & 2478 & 1817 & $\bm{1708}$ \textbf{(2nd)} & 1754 & 2246 & 4093 & $\bm{1674}$ \textbf{(1st)} \\
         & $R^2$ & 0.831 & 0.864 & 0.859 & 0.927 & $\bm{0.939}$ \textbf{(2nd)} & 0.936 & 0.874 & 0.584 & $\bm{0.942}$ \textbf{(1st)} \\
        \hline
    \end{tabular}
    }.    
    \label{tab:market_performance_summary}
\end{table*}

\subsubsection{Forecast Performance Analysis} \label{subsubsec:exp_results_performance}
With all models incorporating mobility features, the comprehensive results are summarized in Table \ref{tab:market_performance_summary}. Note that key benchmarks used in our experiments are extensively tuned, and the results shown in Table \ref{tab:market_performance_summary} reflect the best-performing model with optimal hyperparameter configurations. The details of hyperparameter search are reported in \ref{appendix:hyperparameter_search}.

Notably, the PureKAN model (which aims to assess the applicability of KAN without modifications) consistently presents significantly poorer performance across all three markets compared to the other deep learning benchmarks and the developed LoadKAN. For instance, in the NYISO market, PureKAN yields a MAPE of $9.87\%$ and a low R$^2$ score of $0.281$, indicating its predictions are substantially less accurate than more established architectures. Similarly, high errors are observed in CAISO and ERCOT, with MAPEs of $6.47\%$ and $7.82\%$, and RMSEs of $1943$ MW and $4093$ MW, respectively. These results empirically indicate the challenges that KAN faces in directly modeling complex temporal dependencies without specific architectural adaptations for sequence processing, such as the temporal attention mechanism employed in our LoadKAN. As noted in Section \ref{subsec:intro_related_work}, this finding aligns with discussions in the literature \cite{Jiang2025,Abbas2025} regarding the limitations of standard KAN for direct application to time-series data. Given this considerably weak performance, PureKAN will not be included in the subsequent analysis.

Regarding our developed LoadKAN, the experimental results demonstrate that it remains highly competitive and achieves the best overall balance (i.e., prediction accuracy and transparency) across all three markets. While the extensive hyperparameter tuning enables specific state-of-the-art architectures, such as Informer, to slightly exceed LoadKAN in particular scenarios, our model consistently performs within the top tier while providing the unique advantage of intrinsic interpretability.

\paragraph{NYISO} In the NYISO market, LoadKAN achieves an MAPE of $3.34\%$, an RMSE of $847$ MW, and an R$^2$ score of $0.914$. Under this rigorous hyperparameter search, the Informer model slightly exceeds LoadKAN's predictive accuracy, recording a MAPE of $3.17\%$, an RMSE of $812$ MW, and an R$^2$ of $0.924$. While the Informer provides a marginal accuracy gain, LoadKAN's performance represents a substantial improvement over traditional benchmarks like the MLP ($4.88\%$ MAPE) and remains within a very narrow $0.17\%$ MAPE margin of the market's best performer.

\paragraph{CAISO} Trends in the CAISO market are similar, where LoadKAN attains a low MAPE of $1.94\%$ and RMSE of $641$ MW, alongside a high R$^2$ of $0.971$6. In this market, the TCN model is the most competitive, slightly outperforming our model with a MAPE of $1.81\%$ and an RMSE of $596$ MW. Despite the TCN's slightly higher precision, LoadKAN demonstrates substantial superiority over other standard benchmarks, such as LSTM/GRU (MAPEs above $3.1\%$), and matches the performance of complex transformer-based architectures while maintaining its structural transparency.

\paragraph{ERCOT} The results in the ERCOT market highlight the particular strength of LoadKAN, where it secures the best overall performance with a MAPE of $2.91\%$, RMSE of $1674$ MW, and a leading R$^2$ of $0.942$9. In this market, LoadKAN outperforms the best-performing baseline (Transformer at $2.97\%$ MAPE). The gap between LoadKAN and traditional recurrent models like LSTM and GRU is particularly pronounced, with LoadKAN reducing the MAPE by approximately $29.2\%$ compared to the LSTM's $4.11\%$. Furthermore, LoadKAN reduces the RMSE by over $770$ MW compared to the best recurrent benchmark, indicating its superior ability to capture the high-variance load patterns characteristic of the ERCOT market.

These findings across the three markets demonstrate the effectiveness and competitiveness of the LoadKAN architecture. We explicitly note that LoadKAN does not uniformly outperform all baselines across every market and metric. Specifically, through extensive hyperparameter tuning, the Informer and TCN models achieved marginally higher accuracy in the NYISO and CAISO markets, respectively. However, LoadKAN secures best-in-class results in ERCOT and consistently remains within the top tier (e.g., within 0.2\% in MAPE of the best-performing models) across the others. This indicates that LoadKAN's underlying mechanism, combining dedicated temporal feature processing with KAN's expressive capacity, provides a highly competitive alternative that successfully bridges the gap between predictive power and essential structural interpretability.

Additionally, the time costs for LoadKAN training are $201$, $55$, and $186$ seconds for the NYISO, CAISO, and ERCOT. Notably, in the CAISO market, LoadKAN demonstrates better convergence efficiency, completing training in 55 seconds (faster than both the Transformer and PureKAN). Thus, the marginal computational overhead is a reasonable and necessary trade-off to achieve the framework's core objective: unlike the faster but opaque benchmarks, LoadKAN provides granular interpretability, allowing stakeholders to explicitly visualize and understand the non-linear relationship between each specific feature and the electricity load.

\begin{remark}[Potential Rationales for LoadKAN's Performance]

    In high-dimensional time-series forecasting, models that do not restrict feature interaction at early stages of the neural pipeline may suffer from two issues, as observed and discussed in the literature: 1) Signal Entanglement: In models like transformers, different features are mixed into latent embeddings, potentially leading to signal dominance where a high-variance feature masks the subtle but important temporal signatures of others \cite{NEURIPS2022_91a85f3f,an2025frednspectraldisentanglementtime,10.1145/3534678.3539140,liu2024disentsdisentangledchannelevolving}; 2) Noise Propagation: Full interaction allows noise in one feature to garble the representation of another \cite{NEURIPS2022_91a85f3f}.

    Additionally, in the context of electricity load forecasting, the total observed load can be considered the result of additive superposition of various demand drivers. For instance, the demand surge caused by increased transit mobility and the demand shift caused by residential activity act as distinct signals that sum at the grid level. Benefiting from the additive property of KAN, our LoadKAN represents the model as a collection of these feature-wise contributions, thereby aligning the model structure with the physical law of superposition in power systems.
\end{remark}

\subsubsection{Ablation Study -- The Impact of Mobility Features} \label{subsubsec:exp_results_impact_mobility}
\begin{figure}[!t]
    \centering
    \includegraphics[width=.8\linewidth]{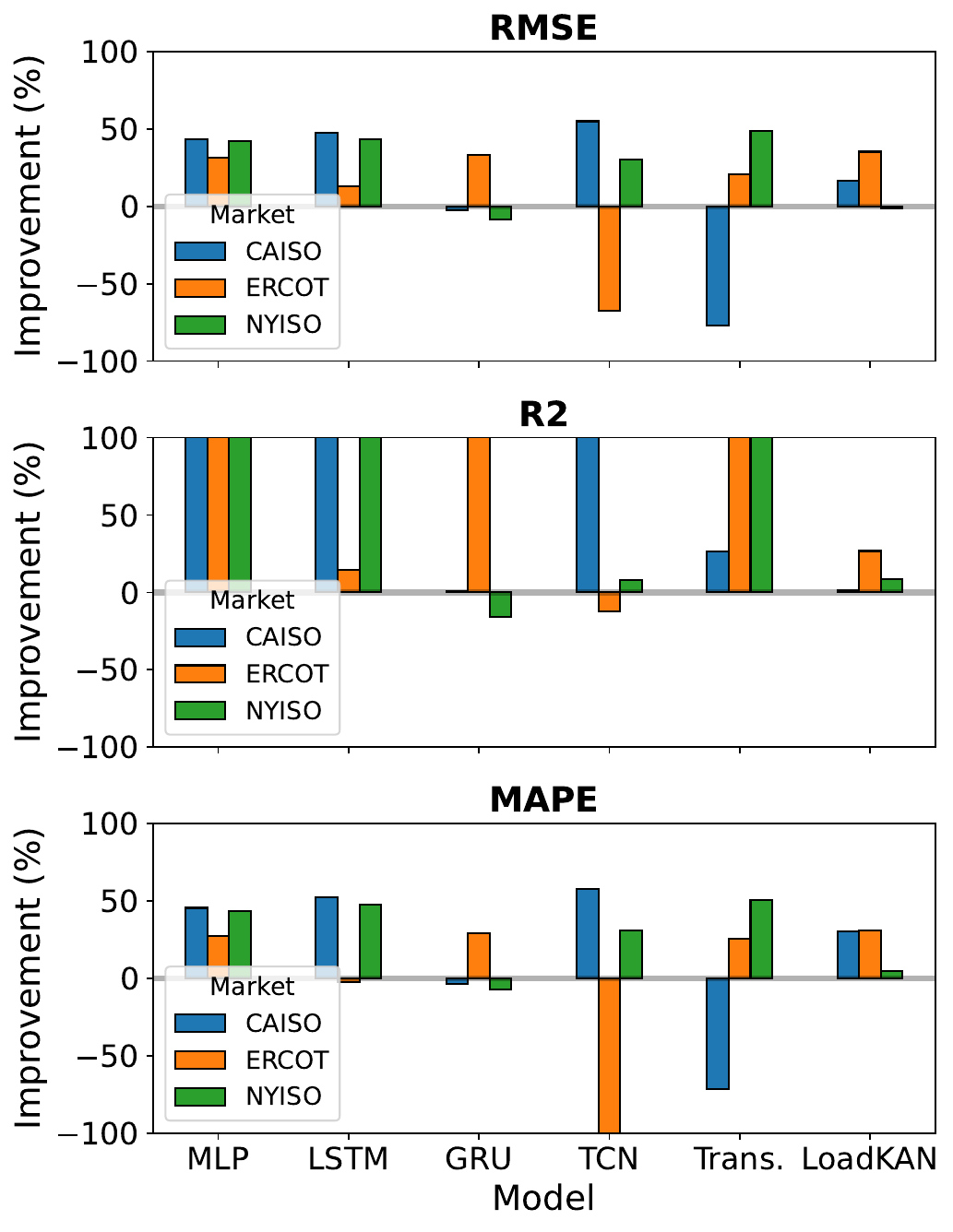}
    \caption{Performance improvements after integrating mobility features.}
    \label{fig:performance_improve_mobility}
\end{figure}

Building upon the demonstrated forecast performance where mobility features are integral, this subsection further assesses their specific contribution through an ablation study, which compares the performance of each model when trained with and without integrating these mobility-related inputs across the NYISO, CAISO, and ERCOT markets. The percentage changes in MAPE, RMSE, and R$^2$ due to the inclusion of mobility features are depicted in Fig.~\ref{fig:performance_improve_mobility}. Note that the improvement percentages have been capped within the range of $[-100\%, 100\%]$ for better readability.

The inclusion of mobility features demonstrates a significantly positive impact on load forecasting performance. Out of $54$ evaluated scenarios, i.e., model-market-metric combinations, approximately $77.8\%$ of scenarios show performance gains. On average, mobility features contribute to an approximate $16.28\%$ reduction in MAPE, a $16.90\%$ reduction in RMSE, and a $189.10\%$ increase in R$^2$ across all combinations. This suggests that mobility data provides valuable contextual signals, capturing dynamic shifts in electricity consumption patterns.

More specifically, integrating mobility data enhances the performance of all neural models, albeit to varying degrees. The developed LoadKAN model consistently and substantially benefits from such data augmentation, achieving notable error reductions such as a $31.15\%$ decrease in MAPE and a $35.32\%$ decrease in RMSE in the ERCOT market, indicating its proficiency in leveraging exogenous features. Other advanced architectures like transformer and TCN also show strong improvements, with the transformer achieving up to a $50.57\%$ MAPE reduction in NYISO, and TCN a $57.75\%$ MAPE reduction in CAISO. Standard sequential models including LSTM and GRU, as well as the simpler MLP, also demonstrate considerable performance gains, exemplified by LSTM's $52.55\%$ MAPE reduction in CAISO and MLP's $45.61\%$ MAPE reduction in the same market, indicating the broad utility of mobility data across different model complexities. These model-specific variations suggest that architectures with greater capacity or more sophisticated mechanisms for feature interaction, such as those in LoadKAN and transformer models, may be better equipped to extract predictive value from diverse data sources like mobility.

The influence of mobility features also reveals some dependency on the specific characteristics of each electricity market. In NYISO, for example, the transformer model's MAPE is reduced by $50.57\%$ and LSTM's R$^2$ increased by a significant $611.75\%$ with mobility data. Similarly, CAISO sees substantial enhancements, such as a $57.75\%$ MAPE reduction for the TCN model and a $118.67\%$ increase in R$^2$ for MLP. In ERCOT, the KAN model's MAPE improves by $31.15\%$, while MLP's R$^2$ sees an increase of $524.14\%$. These regional differences in impact likely stem from unique correlations between population movement and energy usage, driven by factors such as urban density, economic activities, and local commuting habits.

In conclusion, this ablation study confirms that the specified mobility features are a valuable addition to the input feature set for electricity load forecasting. Their integration leads to more accurate predictions across various models and markets. The developed LoadKAN architecture, in particular, demonstrates a strong aptitude for capitalizing on this supplementary information, which contributes significantly to its outstanding forecasting capabilities documented in the overall performance comparison.

\subsubsection{Interpretable Analysis via LoadKAN} \label{subsubsec:exp_results_interpretable} 

In addition to the previous findings of LoadKAN's accurate forecasts and the significant positive impact of mobility features, we now focus on analyzing the model's interpretability to understand how it leverages mobility information. Through the unique architecture of KAN, where activation functions are learnable splines on the edges, this study explores the learned relationships between the six distinct human mobility features and the forecasted electricity load across NYISO, CAISO, and ERCOT. As described in Section \ref{subsec:data}, the mobility features include: 1) Retail \& Recreation; 2) Grocery \& Pharmacy; 3) Parks; 4) Transit Stations; 5) Workplaces; and 6) Residential. The analysis focuses on visualizing and quantifying these relationships through learned activation functions and their sensitivities, aiming to demonstrate LoadKAN's capability to provide not only accurate forecasts but also interpretable insights into the complex interplay between human mobility and electricity consumption.

Our analysis first examines the learned activation functions by the KAN plugin for three-day-ahead electricity load forecasting by using KAN's input (i.e., the output of the feature-isolated attention) related to the six human mobility features. Figs. \ref{fig:learned_activation_func_NYISO}, \ref{fig:learned_activation_func_CAISO}, and \ref{fig:learned_activation_func_ERCOT} depict these activation functions for the three markets, respectively, illustrating the learned relationship between each processed mobility feature and the KAN-output predictions. The significant positive correlations between the processed mobility representation and the corresponding raw mobility input are detailed in \ref{appendix:relation_mobility_representation}, making the scalar representation a high-fidelity proxy for the actual mobility features. Therefore, an increased value of the scalar representation signifies more human presence or activity in that category. For instance, an increased residential value indicates that more people are at their places of residence.

\subparagraph{NYISO} (Fig. \ref{fig:learned_activation_func_NYISO}) operates in a region with high population density, significant commercial activity, and heavy reliance on public transit, especially in New York City. Its climate features distinct seasons with substantial heating and cooling demands.
\begin{itemize}

    \item \textbf{Retail \& Recreation:} Increased activity consistently leads to a slight to moderate positive KAN output across the three days, aligning with higher load from commercial activities at venues such as restaurants and theaters.

    \item \textbf{Grocery \& Pharmacy:} This feature shows a relatively modest impact. Increased mobility towards these essential stores correlates with a slight positive KAN output on Day 2 and Day 3, while the impact becomes negligible or even slightly negative at higher feature values on Day 1.

    \item \textbf{Parks:} A notable positive KAN output is observed with increased mobility to parks and public gardens across all three days. This pronounced effect could reflect energy use in associated facilities (e.g., visitor centers and lighting) within larger parks and energy consumption related to travel to and from these locations.

    \begin{figure*}[!t]
        \centering
        \includegraphics[width=2\columnwidth]{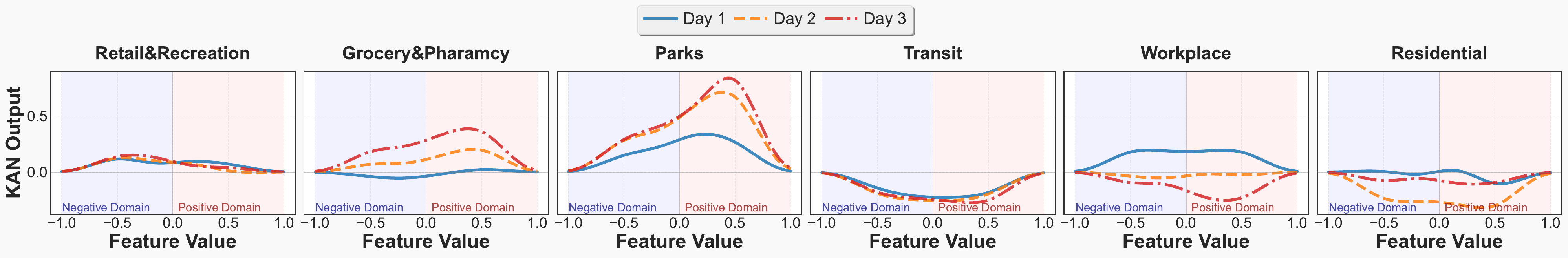}
        \caption{Visualizations of learned activation functions in the NYISO market.}
        \label{fig:learned_activation_func_NYISO}
    \end{figure*}

    \item \textbf{Transit Stations:} Increased transit activity correlates with a consistently negative KAN output across all three days, the magnitude of which is notable. This suggests that greater transit usage may be associated with reduced overall electricity consumption, possibly due to substitution effects, e.g., people leaving more energy-intensive homes or commercial areas.

    \item \textbf{Workplaces:} Higher workplace presence correlates with a notable positive KAN output on Day 1, reflecting increased load from commercial and industrial activity. However, this effect diminishes significantly on Day 2 and Day 3, where the relationship becomes much weaker and the KAN output remains close to zero or may even turn slightly negative at higher feature values -- suggesting a delayed or non-linear response of system load to workplace mobility over time.

    \item \textbf{Residential:} An increase in the residential feature value — indicating more people staying at home — generally leads to a more negative KAN output. This effect is minimal or negligible on Day 1 but becomes more pronounced with a clear negative correlation on Day 2 and Day 3. This pattern implies that increased residential presence may coincide with reduced total system demand, potentially due to concurrent reductions in workplace and commercial activity.

\end{itemize}

\subparagraph{CAISO} (Fig. \ref{fig:learned_activation_func_CAISO}) manages California's grid, characterized by large, dispersed urban areas, significant agricultural and technological sectors, and high renewable penetration, in particular with the increasing installation of solar photovoltaics.
\begin{itemize}
    \item \textbf{Retail \& Recreation:} The relationship between this mobility feature and KAN output is complex. Increased activity (positive feature domain) generally leads to a positive KAN output, typically reaching a slight to moderate peak; however, Day 2 and Day 3 exhibit S-shaped curves where output may initially dip or stay near zero before rising. Conversely, decreased activity (negative feature domain) results in a slightly negative KAN output on Day 1, and more notable negative dips on Day 2 and Day 3 before the output trends back towards zero.

    \item \textbf{Grocery \& Pharmacy:} This feature's impact on KAN output varies significantly across days and activity levels. Increased activity yields a notable positive peak on Days 2 and 3, but presents a mixed effect on Day 1 -- initially positive then slightly negative. Decreased activity leads to a negative dip on Day 1, an initial positive KAN output peak followed by a dip on Day 2, and a very strong positive peak on Day 3.

     \begin{figure*}[!t]
        \centering
        \includegraphics[width=2\columnwidth]{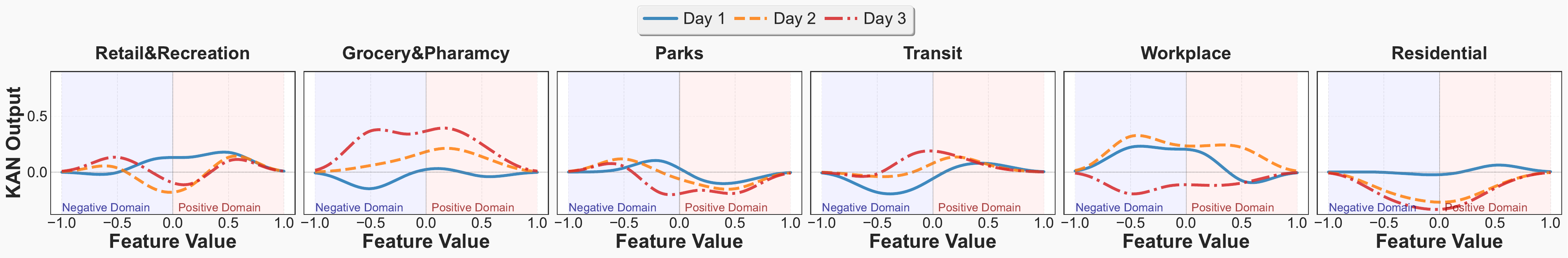}
        \caption{Visualizations of learned activation functions in the CAISO market.}
        \label{fig:learned_activation_func_CAISO}
    \end{figure*}

    \item \textbf{Parks:} The influence of park-related mobility is multifaceted. Increased activity often results in negative or mixed KAN output; Day 1 is predominantly negative, while Day 2 and Day 3 show complex shapes with significant negative dips before potentially turning slightly positive only at higher activity levels. In contrast, decreased activity consistently yields a positive KAN output, with discernible positive peaks of varying intensity across the three days.

    \item \textbf{Transit Stations:} Increased transit activity generally correlates with a positive KAN output across all three days, with moderate positive peaks observed. This positive correlation for increased activity contrasts with the pattern observed in NYISO. For decreased activity, the KAN output is negative on Day 1 and Day 2, but surprisingly shows a strong positive peak on Day 3.

    \item \textbf{Workplaces:} Workplace mobility presents a highly variable impact on KAN output. Increased activity shows a slight positive peak on Day 2, a mixed effect (i.e., initially positive then negative) on Day 1, and is predominantly negative on Day 3. Conversely, and consistently across all three days (except Day 3), decreased activity results in a strong positive KAN output, with substantial positive peaks of varying intensity observed.

    \item \textbf{Residential:} Residential mobility's influence on KAN output is complex and day-dependent. Increased presence leads to a slight positive KAN output on Day 1, but results in a significantly negative output on Days 2 and 3. Decreased presence yields a slightly positive output on Day 1, a significant negative dip on Days 2 and 3.
\end{itemize}

\subparagraph{ERCOT} (Fig. \ref{fig:learned_activation_func_ERCOT}) manages most of Texas's grid, which is largely independent and operates in a hot climate with high air-conditioning load and substantial industrial activity. Urban areas are sprawling with high reliance on personal vehicles.
\begin{itemize}
    \item \textbf{Retail \& Recreation:} Increased activity results in a positive KAN output, though the shape of this response can be complex, with S-curves observed on Day 2 and Day 3 where output may dip before peaking. Decreased activity often shows a mildly positive or mixed (e.g., Day 2) KAN output, with Day 3 exhibiting a clearer positive peak in this domain.

    \begin{figure*}[!t]
        \centering
        \includegraphics[width=2\columnwidth]{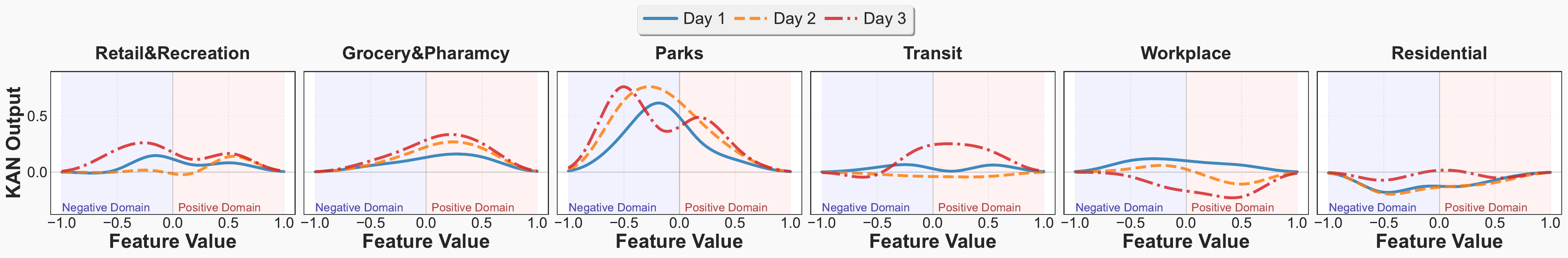}
        \caption{Visualizations of learned activation functions in the ERCOT market.}
        \label{fig:learned_activation_func_ERCOT}
    \end{figure*}

    \item \textbf{Grocery \& Pharmacy:} Increased activity consistently leads to a noticeable positive KAN output across all three days. When activity decreases, all three days exhibit a clear positive KAN output in this domain.

    \item \textbf{Parks:} This feature demonstrates a very substantial KAN output, which is particularly pronounced when park visitation decreases; here, extremely strong positive peaks are evident on all three days, often dominating the KAN response. When park visitation increases, there is also a significant positive KAN output forming notable peaks, though these are generally less pronounced than the peaks observed in the negative domain. The shape of the curve in the positive domain can also include dips after an initial peak, as seen on Day 3.

    \item \textbf{Transit Stations:} The relationship between transit station activity and KAN output is complex and varies daily. Increased activity leads to a slightly positive KAN output on Day 1, is mostly flat or slightly negative on Day 2, and shows a clearer positive peak on Day 3 before dipping. Decreased activity results in a slightly negative output on Day 1, is flat near zero on Day 2, and shows a notable positive peak on Day 3.

    \item \textbf{Workplaces:} The impact of workplace presence on KAN output is highly variable and not consistently positive with increased activity. Increased presence leads to a positive KAN output on Day 1, a mixed response (dipping negative then becoming positive) on Day 2, and a noticeable negative output on Day 3. Decreased presence also shows complex behavior, with slightly positive KAN outputs on Day 1 and 2, while a notable negative response appears on Day 3.

    \item \textbf{Residential:} Both decreased and increased activities correspond to negative KAN output for this mobility category.
\end{itemize}

\paragraph{Quantitative Sensitivity Analysis of Mobility Features}
To further quantify the influence of each mobility feature, we analyze the sensitivity of the KAN output with respect to changes in each feature. Sensitivity, in this context, is defined as the derivative of the learned activation functions, indicating the rate at which the KAN output changes in response to a change in the input feature. Fig. \ref{fig:sensitivity_NYISO}, \ref{fig:sensitivity_CAISO}, and \ref{fig:sensitivity_ERCOT} present the average absolute sensitivity and average maximum absolute sensitivity for each feature across both the three markets and the forecast horizon. These metrics help identify which features induce the most significant changes in load predictions for given changes in their values.

\begin{itemize}
    \item \textbf{NYISO Sensitivity} (Fig. \ref{fig:sensitivity_NYISO}): The Parks feature exhibits the highest sensitivity by a significant margin, with an average absolute sensitivity of approximately $0.615$ and an average maximum sensitivity of $1.543$. Transit Stations, Grocery \& Pharmacy, and Residential show moderate sensitivities. Workplaces and Retail \& Recreation are the least sensitive on average. This high sensitivity for Parks aligns with its activation function depicted in Fig. \ref{fig:learned_activation_func_NYISO}, which shows a pronounced and relatively steep change in KAN output with increased park visitation. This suggests that factors correlated with park visitation -- such as specific weather conditions, energy use for travel to/from parks, or activities in associated recreational facilities -- have a highly dynamic impact on load. The moderate sensitivity of Transit Stations and Residential reflects the clear and responsive shapes of their activation functions -- consistently negative for Transit, and increasingly negative for Residential on Days 2 and 3. Conversely, the lower average sensitivity for Workplaces might reflect its strong impact being concentrated on Day 1 before diminishing, while for Retail \& Recreation, it aligns with its slight to moderate impact, suggesting their influence, while definite, might be characterized by more gradual average slopes or less sustained steepness across the entire input domain and forecast horizon.

    \begin{figure}[!t]
        \centering
        \includegraphics[width=\columnwidth]{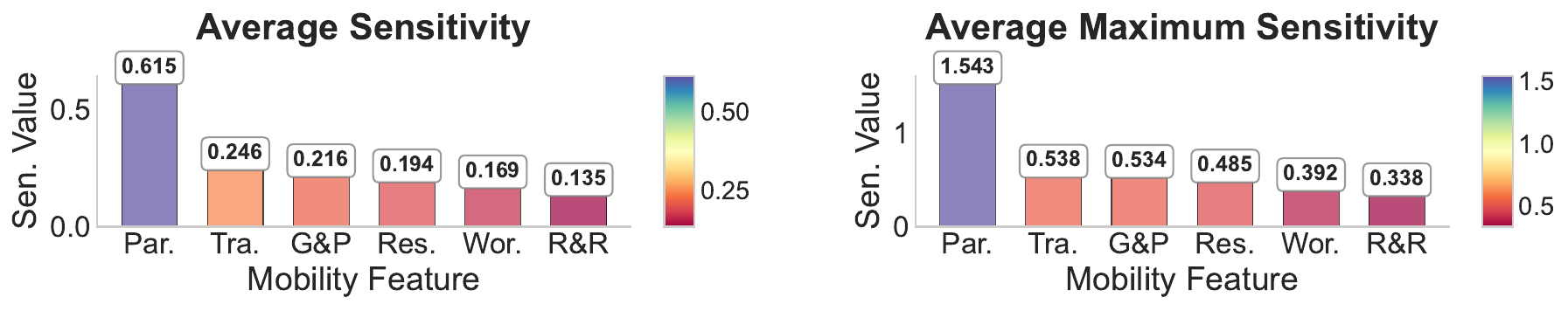}
        \caption{Sensitivity of learned activation functions in the NYISO market.}
        \label{fig:sensitivity_NYISO}
    \end{figure}
        
    \item \textbf{CAISO Sensitivity} (Fig. \ref{fig:sensitivity_CAISO}): In CAISO, the sensitivity is more evenly distributed compared to NYISO and ERCOT. Retail \& Recreation shows the highest average sensitivity ($0.307$), closely followed by Grocery \& Pharmacy ($0.283$) and Workplaces ($0.280$). Parks has an average sensitivity of approximately $0.257$. For average maximum sensitivity, Workplaces ranks highest ($0.743$), followed by Retail \& Recreation ($0.708$). Residential exhibits the lowest average maximum sensitivity ($0.415$). The high sensitivity of commercial features, including Retail \& Recreation, Workplaces, and Grocery \& Pharmacy, is consistent with their activation functions shown in Fig. \ref{fig:learned_activation_func_CAISO}, which outline complex and dynamic changes, including significant peaks, dips, and S-curves, indicating responsive KAN outputs. The lower average maximum sensitivity for Residential, despite its complex activation function with notable dips on Days 2 and 3, suggests that while the relationship is intricate, the steepest slopes within its learned relationship might be less extreme on average than those for the leading commercial features.

    \begin{figure}[!t]
        \centering
        \includegraphics[width=\columnwidth]{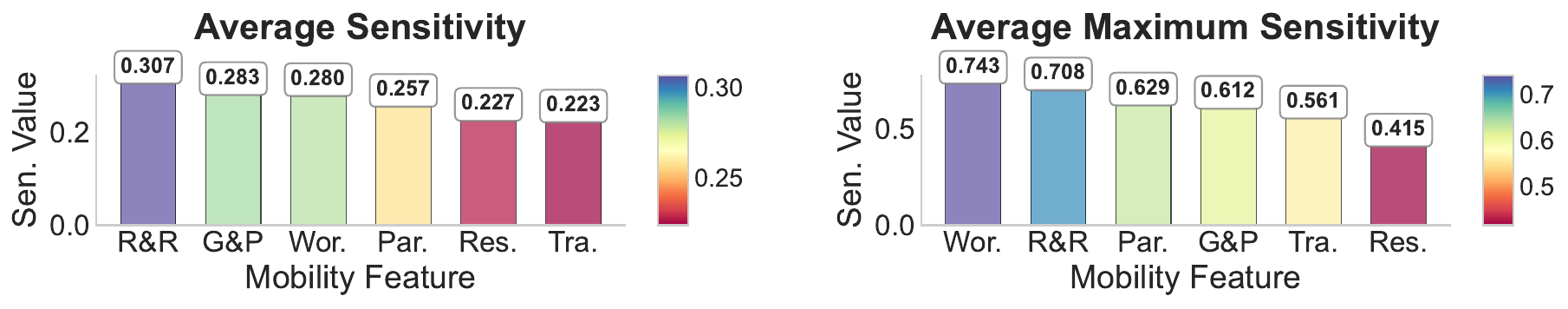}
        \caption{Sensitivity of learned activation functions in the CAISO market.}
        \label{fig:sensitivity_CAISO}
    \end{figure}

    \item \textbf{ERCOT Sensitivity} (Fig. \ref{fig:sensitivity_ERCOT}): Similar to NYISO, the Parks feature overwhelmingly dominates in ERCOT, with the highest average absolute sensitivity ($0.743$) and average maximum sensitivity ($1.682$). Other features like Grocery \& Pharmacy (average sensitivity $0.251$) and Retail \& Recreation ($0.222$) show moderate average sensitivities. Transit Stations presents the lowest average sensitivity ($0.152$) in ERCOT. This exceptional sensitivity for Parks directly corresponds to its activation function illustrated in Fig. \ref{fig:learned_activation_func_ERCOT}, which shows very substantial KAN outputs, including extremely strong positive peaks when park visitation decreases and significant positive peaks when it increases, and also exhibits steep slopes indicating a strong and rapid response of load to changes in park mobility. This suggests that activities or conditions strongly correlated with park mobility (e.g., energy use in recreational facilities, energy use during periods of high outdoor leisure activity which might also drive HVAC loads) significantly influence load. The lower sensitivity of Transit Stations is also reflected in its more subdued and complex activation function shape in ERCOT, which reflects a less direct and more varied impact compared to NYISO.
    
    \begin{figure}[!t]
        \centering
        \includegraphics[width=\columnwidth]{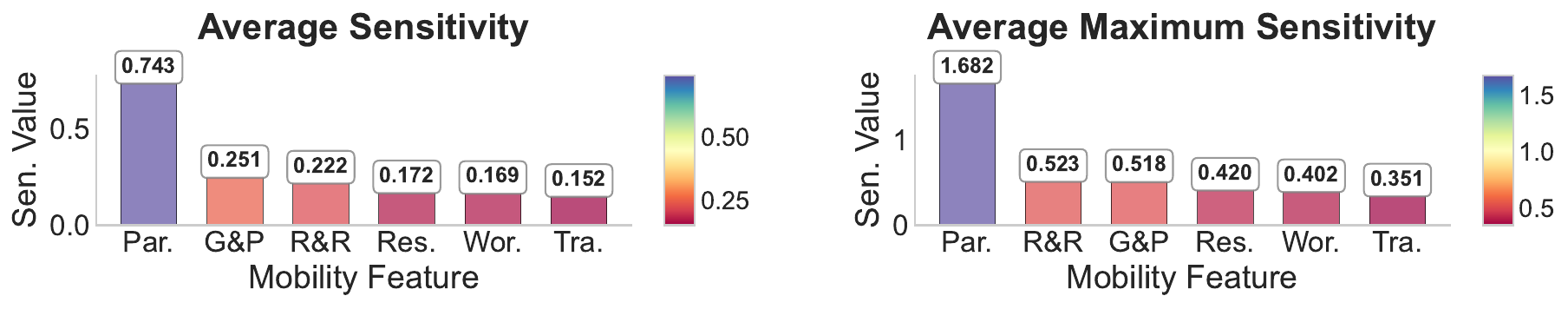}
        \caption{Sensitivity of learned activation functions in the ERCOT market.}
        \label{fig:sensitivity_ERCOT}
    \end{figure}
        
\end{itemize}
Comparing across markets, the Parks feature is exceptionally sensitive in NYISO and ERCOT, a finding that resonates with the strong and often steep responses observed in their activation functions shown in Fig. \ref{fig:learned_activation_func_NYISO} and \ref{fig:learned_activation_func_ERCOT}, respectively. This insight suggests that factors associated with mobility to parks are highly dynamic drivers of load in these regions. In contrast, CAISO shows higher sensitivity for commercial-activity-related features such as Workplaces, Retail \& Recreation, and Grocery \& Pharmacy, which aligns with the complex and responsive shapes of their activation functions in Fig. \ref{fig:learned_activation_func_CAISO}. The Transit Stations feature's moderate sensitivity in NYISO, compared to its lower sensitivity in ERCOT and CAISO, mirrors the notable and consistent (though negative) impact seen in its NYISO activation function. This sensitivity analysis, therefore, provides a quantitative measure of feature influence that not only ranks features by their rate of impact but also complements and reinforces the qualitative interpretations of the activation function shapes, highlighting which features the model deems most influential in driving load changes.

\subparagraph{Similarities Across Markets}
Despite regional differences, several common trends emerge when analyzing the learned relationships between mobility features and electricity load, as captured by LoadKAN's activation functions and sensitivities:
\begin{itemize}
    \item \textbf{Workplace Activity Impact:} Increased presence at workplaces often shows a positive KAN output, particularly on Day 1 in NYISO and ERCOT, reflecting initial increases in commercial and industrial electricity consumption. However, this effect is not consistently positive across all markets or all forecast days. In NYISO, the positive impact diminishes significantly on Day 2 and Day 3. In CAISO, increased activity has a mixed effect on Day 1, a slight positive peak on Day 2, and is predominantly negative on Day 3. In ERCOT, after a positive Day 1, the response is mixed on Day 2 and negative on Day 3. Decreased workplace activity in CAISO, notably, can lead to a strong positive KAN output. This highlights a complex, time-dependent, and market-specific relationship rather than a uniformly positive correlation.
    
    \item \textbf{Retail \& Recreation Activity Impact:} Increased mobility towards Retail \& Recreation locations leads to a positive KAN output, corresponding to energy use in these commercial sectors. This is observed with slight to moderate positive output in NYISO. In CAISO and ERCOT, while positive for increased activity, the relationship can be more complex and is often featured by S-shaped curves or initial dips before rising, especially on Day 2 and Day 3.
    
    \item \textbf{Residential Presence Impact:} The impact of an increase in the Residential feature value (i.e., more people at home) on KAN output varies significantly and is often negative, contrary to simple intuitions of increased home energy use directly leading to higher overall system load. In NYISO, increased residential presence leads to a more negative KAN output, especially on Day 2 and Day 3. In CAISO, it leads to a slight positive KAN output on Day 1 but a significantly negative output on Day 2 and Day 3. In ERCOT, both increased and decreased residential activity correspond to a negative KAN output. These patterns suggest that increased residential presence often coincides with reductions in other potentially more energy-intensive activities (e.g., at workplaces or commercial areas), leading to a net decrease in the KAN model's system-wide load prediction. The specific shapes and magnitudes vary, reflecting diverse regional factors.
\end{itemize}

\subparagraph{Significant Differences Across Markets}
The learned relationships and their sensitivities also highlight distinct market-specific characteristics:
\begin{itemize}
    \item \textbf{Dominance and Nature of Parks Impact:} The Parks feature reflects strikingly high KAN output and sensitivity in ERCOT and NYISO. In ERCOT, such effect is particularly pronounced, with extremely strong positive KAN output peaks when park visitation \textit{decreases}, alongside significant positive outputs when it increases. In NYISO, increased park visitation leads to a notable positive KAN output. This suggests factors correlated with park mobility are highly impactful and responsive load drivers in these regions. In CAISO, while mobility to Parks influences load (e.g., increased activity often yielding negative or mixed output, and decreased activity yielding positive output), its sensitivity and overall KAN output effect are less dominant compared to its commercial features.

    \item \textbf{Role of Transit Stations Activity:} The Transit Stations feature demonstrates a notable impact and moderate sensitivity in NYISO, where increased activity consistently correlates with a negative KAN output, contrasting with CAISO where it correlates with a positive output. This directly aligns with New York's greater reliance on public transportation, where changes in transit use appear more strongly indicative of broader load-affecting activity shifts (e.g., people leaving energy-intensive homes or commercial areas). In ERCOT, its lower sensitivity, along with complex and varied activation shape, reflects its lesser and more ambiguous role in overall load dynamics.
    
    \item \textbf{Variations in Commercial Feature Sensitivities:} While mobility to commercial locations influences load, their relative sensitivities differ. CAISO, for instance, shows a cluster of high sensitivity among these features, whereas in NYISO and ERCOT, the Parks feature overshadows them in terms of sensitivity.
    
    \item \textbf{Complexity of Activation Function Shapes and Sensitivity Magnitudes:} Beyond general trends, the exact shapes of activation functions (e.g., presence of multiple peaks, dips, S-curves, saturation points) and particularly their sensitivity magnitudes differ markedly. For instance, the Residential feature's activation function in CAISO exhibits complex day-dependent behavior, i.e., slight positive output on Day 1, significant negative output/dips on Days 2 and 3 for increased presence, yet its average maximum sensitivity is the lowest among all features in that market, contrasting with its moderate sensitivity ranking in NYISO where increased presence also leads to a negative KAN output, especially Day 2-3. This implies different behavioral thresholds and response intensities.

    \item \textbf{Temporal Evolution in Activation Functions:} The way activation functions for specific features evolve from Day 1 to Day 3 predictions differs among markets, as seen in Fig. \ref{fig:learned_activation_func_CAISO} and \ref{fig:learned_activation_func_ERCOT}. This indicates that the model learns varying short-term versus slightly longer-term impacts of mobility patterns. For example, in ERCOT, the KAN output for increased mobility to Parks shows changing shapes (e.g., dips after initial peak on Day 3), while for decreased mobility, the extremely strong positive peaks are evident across all three days. For Transit Stations in ERCOT, the impact varies from slightly positive on Day 1, to flat/slightly negative on Day 2, to a positive peak then dip on Day 3. In CAISO, the Residential function not only becomes more complex but also shows a shift from slightly positive KAN output for increased presence on Day 1 to significantly negative on Days 2 and 3.
\end{itemize}

\subparagraph{Interpretation Based on Market Characteristics}
The observed differences and similarities in activation functions and their sensitivities can be largely attributed to the unique socio-economic fabric, climate, and infrastructure of each electricity market's jurisdictions:
\begin{itemize}
    \item \textbf{NYISO:} The dense urban environment and high public transit usage are reflected in the notable negative impact and moderate sensitivity of Transit Stations activity, possibly indicating a shift away from more energy-intensive locations. The moderate impact of Workplaces activity on the load is also characteristic. The high sensitivity and positive impact associated with Parks visitation might reflect energy consumption linked to specific types of recreational activities or facilities, travel, or could be a proxy for weather conditions. The distinct seasons likely make residential load patterns complex, which showed a negative KAN output with increased presence, especially on later days, possibly reflecting a greater reduction in commercial/industrial load when more people are home.

    \item \textbf{CAISO:} California's diverse geography, sprawling urban centers, and distinct economic activities likely contribute to the complex activation function shapes and the higher sensitivity of commercial features like Workplaces, Retail \& Recreation, and Grocery \& Pharmacy. Significant cooling demand, varied recreational patterns, and high renewable penetration could all influence these intricate relationships.

    \item \textbf{ERCOT:} The hot climate and large industrial base are key influencers. The exceptionally strong KAN output and highest sensitivity associated with Parks mobility could be linked to energy use in recreational facilities, or it may strongly correlate with widespread residential/commercial air conditioning use during periods of outdoor leisure activity (or lack thereof, prompting other activities). The Workplaces signal indicates a complex industrial/commercial response over the forecast horizon. The lower sensitivity of Transit Stations is consistent with Texas's predominantly car-centric transportation. The consistently negative KAN output for Residential mobility suggests complex interactions with other consumption sectors or baseline load patterns in this market.
\end{itemize}

To further demonstrate the KAN's interpretability, in addition to the above quantitative analysis, we further calculate feature importance scores and correlations between learned spline gradients and load changes. Detailed results are reported in \mbox{\ref{appendix:feature_importance_gradient_correlation}}. The findings are consistent with both spline function visualizations and sensitivity analysis. Note that our study does not treat the model's interpretability as a mathematical prerequisite for the provided sensitivity analysis. Rather, the sensitivity analysis serves as a secondary, independent validation to quantify the relative significance of relationships (identified by KAN) across the three electricity markets.

In summary, this detailed interpretable analysis, combining the qualitative insights from activation function shapes with quantitative measures, demonstrates that the LoadKAN model, to a great extent, learns and elucidates market-specific, non-linear relationships between human mobility patterns and electricity load. The variations across NYISO, CAISO, and ERCOT suggest the importance of considering unique regional socio-economic characteristics, climate conditions, and infrastructure in load forecasting models. These insights highlight how different mobility aspects drive load changes with varying intensity and responsiveness, offering a deeper understanding beyond mere forecast accuracy.

\subsubsection{Discussion on Relations between Human Mobility and Electricity Load} \label{subsubsec:exp_results_discussion} 
The above detailed interpretable analysis via LoadKAN, consisting of both the shapes of learned activation functions and their quantitative sensitivities, reveals a multifaceted and regionally distinct relationship between human mobility patterns and electricity load. The following discussion synthesizes these findings from Section \ref{subsubsec:exp_results_interpretable} to explore the broader implications of these observed dependencies.

A primary observation is that human mobility serves as a significant and notably complex driver of electricity consumption patterns as learned by the model. Increased activity in Retail \& Recreation and Grocery \& Pharmacy points towards increased KAN output, though the specifics vary by market and can involve complex non-linearities like S-curves or dips. The impact of Workplaces mobility is highly variable: often positive on the first day of forecast but diminishing, becoming mixed, or even negative on subsequent days or for increased activity in certain markets. This deviates from a simple assumption that more people at work always means higher system load over a multi-day horizon. Perhaps most strikingly, increased Residential presence frequently correlates with a \textit{negative} KAN output, suggesting that system-wide load may decrease, possibly due to concurrent reductions in commercial or industrial activity outweighing increased domestic use. The Transit Stations feature shows market-dependent impacts: a notable negative KAN output with increased activity in NYISO, contrasting with a positive correlation in CAISO, reflecting differing transportation roles. The Parks feature emerges as surprisingly influential and highly sensitive, especially in ERCOT and NYISO, suggesting that factors associated with park-related mobility are substantial and volatile contributors to the model's load predictions, potentially linked to associated travel, facility use, or correlated weather patterns.

Moreover, the LoadKAN model demonstrates that these relationships are far from simple linear correlations. The learnable spline activation functions capture significant non-linearities, including varying sensitivities across the range of mobility changes, saturation effects, and potential threshold phenomena. For instance, while an increase in a particular mobility feature points to a directional change in KAN output, the \textit{magnitude} and even direction of this change can differ substantially depending on the current level of mobility, the specific market, and the forecast day. The high sensitivity and dramatic KAN output changes related to Parks mobility in ERCOT, for example, imply that load predictions in this region are particularly responsive to fluctuations in this specific type of mobility, more so than to changes in other mobility categories that might otherwise be considered primary load drivers. This highlights the model's ability to uncover non-obvious yet impactful relationships.

The market-specific nature of these relationships, revealed through both activation functions and sensitivities, is a crucial insight. The dominance of commercial activity sensitivities in CAISO, the pronounced impact and sensitivity related to Parks mobility in ERCOT and NYISO, and the notable role and sensitivity of Transit Stations in NYISO highlight how regional demographics, economic structures, urban layouts, transportation infrastructure, and climate dictate the distinct ways human movement translates into electricity demand predictions. For example, the lower sensitivity of Transit Stations in ERCOT compared to NYISO clearly reflects differing reliance on public transportation. Similarly, the intense sensitivity and strong KAN output related to Parks visitation changes in ERCOT could be amplified by its hot climate, where park-related mobility (or lack thereof) might coincide with significant HVAC load shifts or energy use in specific types of recreational facilities.

Understanding these granular, feature-specific, and market-dependent relationships between human mobility and electricity load has significant implications. For load forecasting, incorporating such detailed mobility features and leveraging models like LoadKAN that can capture their complex and non-linear impacts and enhance prediction accuracy, particularly for short to medium-term horizons where mobility fluctuations are prominent. Beyond forecasting, these insights can inform grid management strategies. For instance, anticipating load shifts due to evolving mobility trends, such as changes in work-from-home patterns affecting Workplaces and Residential KAN outputs (which are not always intuitive) or planning for the impact of large public events affecting mobility to Parks or Transit Stations, becomes more data-driven. Furthermore, the varying sensitivities suggest where targeted demand-response initiatives, if linked to anticipated mobility changes, might be most effective.

While this study provides valuable insights, the relationships between human mobility and electricity load are part of a larger and interconnected system. The mobility features themselves, though treated as separate inputs, can have interdependencies in the real world. For example, increased workplace presence often implies decreased residential presence during work hours, and the model learns the net effect of these shifts. Moreover, the impact of external factors not directly included in this mobility-focused feature set, such as concurrent weather conditions, day type (weekday/weekend/holiday), and underlying economic shifts, undoubtedly modulate these relationships. Integrating these additional dimensions could further refine the understanding and predictive power of models like LoadKAN.

It is important to note that the mobility-load relationships learned in this study are derived from pandemic-era data. While this specific regime heavily influenced certain dynamics, such as the pronounced load reduction during increased residential presence, these learned relationships maintain a significant degree of post-pandemic generalizability. Recent literature indicates that post-COVID mobility patterns have not fully reverted to pre-pandemic norms, largely due to the permanent structural entrenchment of hybrid and remote work arrangements \mbox{\cite{postcovidmobility_ref1,postcovidmobility_ref2,postcovidmobility_ref3}}. Consequently, the complex interplay between residential and commercial energy consumption captured by our model remains highly relevant today. Furthermore, LoadKAN successfully captures universal, regime-agnostic temporal rhythms, such as the inherent distinctions between weekday and weekend mobility patterns, demonstrating the robust and enduring applicability of these learned relationships.

In summary, the interpretable framework of LoadKAN, through its combined analysis of activation functions and sensitivities, provides a powerful lens to dissect the complex nexus of human mobility and electricity consumption. It moves beyond simple correlations to reveal nuanced, market-specific, and non-linear dependencies, offering actionable insights for more accurate load forecasting and informed energy system planning. The ability to not only predict but also to understand \textit{why} predictions are made, based on the learned influence of specific real-world features like human mobility, is a crucial step towards more robust and intelligent energy management.

\subsection{Limitations} 
While the LoadKAN framework demonstrates highly competitive accuracy and interpretability across the tested scenarios, we acknowledge the limitations of our study to contextualize the findings.

First, our evaluation is limited to three major U.S. electricity markets (NYISO, CAISO, and ERCOT). While these markets offer diverse climatic and economic profiles, they operate under advanced infrastructure and mature data recording standards. The model's performance on grids with less stable data quality or significantly different consumption behaviors (e.g., in developing regions) remains to be verified.

Second, our ablation study confirms that a significant portion of the accuracy gain stems from the integration of granular human mobility features. Consequently, the model's effectiveness is contingent on the availability of such high-quality external data sources. In scenarios where real-time mobility data is unavailable or restricted by privacy regulations, the performance advantage of the framework may be reduced.

Finally, the feature-isolated temporal attention mechanism introduces an inherent structural trade-off. To preserve the purity of individual feature representations and prevent signal entanglement prior to the interpretable KAN head, the architecture explicitly limits cross-feature interactions during the temporal processing stage. While this design choice is essential for accurately interpreting the isolated impact of individual drivers (such as specific human mobility categories), it restricts the model's capacity to capture complex, synergistic interdependencies between different input variables.

\section{Conclusion and Future Works} \label{sec:conclusion}

This paper develops LoadKAN, a novel hybrid deep learning framework with the integration of KAN for interpretable electricity load forecasting. LoadKAN uniquely designs a feature-isolated temporal attention mechanism, aiming to independently extract robust temporal patterns from each input feature sequence, together with a KAN module for the final interpretable load prediction. The incorporation of KAN effectively transforms the black-box neural forecasting model into an intrinsically interpretable one, enabling analysis of how input features, especially human mobility, influence electricity load. Our comprehensive experiments across three representative U.S. electricity markets -- NYISO, CAISO, and ERCOT -- demonstrate that LoadKAN remains highly competitive when compared to extensively-tuned, state-of-the-art, black-box deep learning benchmarks. Further, we quantitatively affirm the substantial positive impact of incorporating human mobility features in enhancing load forecasting performance via a detailed ablation study. For example, our sensitivity analysis identifies distinct regional drivers, such as the ``Parks'' mobility feature in ERCOT, which exhibits a dominant average sensitivity of 0.743 compared to other input features. More importantly, by leveraging the learnable splines of the KAN module, LoadKAN successfully elucidates complex, market-specific, and non-linear relationships between six distinct human mobility features and electricity load, offering insights into load dynamics that are typically opaque in conventional models. This work highlights the potential of hybrid KAN-based models to advance the load forecasting field by providing tools that are both highly predictive and transparent.

For future work, while LoadKAN models individual feature temporal dynamics effectively, future research could explore methods to explicitly model potential interdependencies and interactions among the different mobility features themselves before or within the KAN layer, thereby providing a more holistic view of behavioral impacts.

\appendix

\section{Benchmark Descriptions and Configurations}

\label{appendix:benchmark_description_and_config}

The descriptions and hyperparameter configurations of benchmarks are provided below.
\begin{itemize}
    \item \textbf{MLP:} A foundational feedforward neural network, serving as a standard baseline in numerous deep learning applications.

    Two hidden layers, each containing $32$ neurons and employing the rectified linear unit (ReLU) activation function. The input sequence is flattened to a vector of size $L \cdot M$ before being fed into the first hidden layer.
    
    \item \textbf{LSTM:} A type of RNN specifically designed to capture long-range temporal dependencies in sequential data~\mbox{\cite{lstm}}.

    It consists of two stacked LSTM layers, each with $32$ hidden LSTM cells. A dropout rate of $0.3$ is applied after each LSTM layer (except the last) to reduce overfitting. The output from the last time step of the final LSTM layer is passed to a final linear layer for forecast.
    
    \item \textbf{GRU:} Another variant of RNN, often considered a computationally efficient alternative to LSTM~\mbox{\cite{gru}}.

    It is configured similarly to the LSTM, with two stacked GRU layers, each having $32$ hidden units. A dropout rate of $0.3$ is applied between GRU layers. The output from the final GRU layer's last time step is used for prediction via a linear layer.

    \item \textbf{TCN:} An architecture applying convolutional layers to sequence data~\mbox{\cite{tcn}}.

    Implemented using two residual blocks of temporal convolution. Each block employed causal convolutions with a kernel size of $3$ and $32$ output channels. Standard TCN practices like weight normalization and ReLU activations are used. A dropout rate of $0.2$ is applied within the residual blocks. The output sequence is processed using average pooling before a final linear mapping.

    \item \textbf{Transformer:} A model based on the self-attention mechanism, effectively modeling global dependencies~\mbox{\cite{transformer}}.

    An encoder-only transformer architecture is used. Key parameters include: embedding dimension of $32$, $4$ attention heads in the multi-head self-attention layers, and $2$ encoder layers. The feed-forward networks within each layer use $64$ units with ReLU activation. A dropout rate of $0.1$ is applied throughout the model.

    \item \textbf{Informer:} A state-of-the-art transformer-based baseline ~\mbox{\cite{zhou2021informer}}.

    Its configuration follows the original implementation.

    \item \textbf{Chronos:} A pre-trained model based on T5 language model architecture \mbox{\cite{ansari2024chronos}}.

    Since Chronos is primarily designed for univariate probabilistic time-series forecasting, we adapt it to the present multivariate daily load forecasting task by using the historical load sequence within the same rolling-window setup as the forecasting context. The model receives the past $L$-day load observations and generates the $H$-step ahead daily load forecast. Therefore, Chronos serves as a pretrained foundation-model benchmark for load-only temporal forecasting, while the other neural benchmarks and LoadKAN use the full multivariate input feature set, including weather, price, and mobility variables. Its configuration follows the original implementation.
    
    \item \textbf{PureKAN:} A baseline model composed of stacked KAN layers~\mbox{\cite{Abbas2025,Jiang2025}}.

    It consists of three stacked KAN layers. Similar to MLP, the input sequence is flattened. Each KAN layer uses $4$ basis functions for its splines with a polynomial degree of $3$.

\end{itemize}
Note that a final fully-connected neural layer, also known as the linear layer, is used in most benchmark architectures, including MLP, LSTM, GRU, TCN, transformer, to map the network's internal representation to the required prediction dimension; our LoadKAN model and the PureKAN model inherently produce the $H$-dimensional output from their final KAN layer.

For our LoadKAN, the output channel number of the input 1D convolution projection is the number of input features (i.e., $M$) multiplied with a hidden dimension (set to $32$ by default). The feature-isolated attention mechanism has $2$ layers and $4$ attention heads per layer. The output channel number of the output 1D convolution projection is set to $32$. The number of basis functions for each spline is set to 4 and the polynomial degree of each spline is set to 3. The output dimension of the KAN layer corresponds to the forecast horizon of $H$.

\section{Detailed Formulation of Forecasting Evaluation Metrics} \label{appendix:evaluation_metrics}

The evaluation metrics aggregate the forecasting performance across all $H$ forecast steps for all initiation times $t\in \mathcal{T}_\text{test}$. Let $N_\text{test}^\text{all} = N_\text{test} \cdot H$ be the total number of individual actual-predicted point pairs $(y_{t+h}, \hat{y}_{t+h})$ within the test set. Let $\bar{y}$ be the mean of all $N_\text{test}^\text{all}$ individual actual load values $y_{t+h}$ across the entire test set, which is defined as $\bar{y} = \frac{1}{N_\text{test}^\text{all}} \sum_{t \in \mathcal{T}_{\textrm{test}}} \sum_{h=1}^{H} y_{t+h}$. The three metrics are calculated over these $N_\text{test}^\text{all}$ individual points as follows.
\begin{align}
\label{eq:metric_RMSE}
    \mathrm{RMSE} &= \sqrt{\frac{1}{N_\text{test}^\text{all}} \sum_{t \in \mathcal{T}_{\textrm{test}}} \sum_{h=1}^{H} (y_{t+h} - \hat{y}_{t+h})^2}, \\
\label{eq:metric_MAPE}
    \mathrm{MAPE} &= \frac{100\%}{N_\text{test}^\text{all}} \sum_{t \in \mathcal{T}_{\textrm{test}}} \sum_{h=1}^{H} \left| \frac{y_{t+h} - \hat{y}_{t+h}}{y_{t+h}} \right|, \\
\label{eq:metric_R2}
    \mathrm{R}^2 &= 1 - \frac{\sum_{t \in \mathcal{T}_{\textrm{test}}} \sum_{h=1}^{H} (y_{t+h} - \hat{y}_{t+h})^2}{\sum_{t \in \mathcal{T}_{\textrm{test}}} \sum_{h=1}^{H} (y_{t+h} - \bar{y})^2}.
\end{align}
These three metrics, while all evaluating forecast accuracy on $\mathcal{D}_{\text{test}}$ by comparing individual predicted points $\hat{y}_{t+h}$ to actual points $y_{t+h}$, offer distinct perspectives on model performance. Both RMSE and MAPE directly quantify prediction errors, with lower values indicating higher accuracy. However, they differ significantly in scale and sensitivity. RMSE expresses the average error magnitude in the original units of the load, i.e., MW, and is heavily influenced by large deviations due to the squaring of errors. In contrast, MAPE provides a scale-independent measure by calculating the average absolute percentage error, making it useful for comparisons across different load scales or time periods, though it can be sensitive to near-zero actual values and is undefined if any $y_{t+h}$ equals zero. The $\mathrm{R}^2$ metric differs from the other two as it is not a direct measure of error magnitude but rather quantifies the goodness-of-fit. It indicates the proportion of variance in the actual load across all $N_\text{test}^\text{all}$ points explained by the forecast, relative to a baseline model that simply predicts the overall mean load $\bar{y}$. An $\mathrm{R}^2$ value closer to 1 indicates a better fit, while values close to 0 or negative suggest the model performs poorly compared to the mean baseline.

\section{Hyperparameter Grid Search for Key Benchmarks} \label{appendix:hyperparameter_search}

We extensively tune all baselines with respect to layer depth, width, structural parameters, and training parameters (e.g., learning rates). The details of refined hyperparameter testing per baseline are presented as follows.
\begin{itemize}

    \item MLP: layers $\in \{1, 2, 3, 4\}$, hidden units $\in \{16, 32, 64, 128\}$.
    
    \item LSTM: layers $\in \{1, 2, 3, 4\}$, dropout $\in \{0, 0.1, 0.2\}$, hidden units $\in \{16, 32, 64, 128\}$.

    \item GRU: layers $\in \{1, 2, 3, 4\}$, dropout $\in \{0, 0.1, 0.2\}$, hidden units $\in \{16, 32, 64, 128\}$.

    \item TCN: layers $\in \{1, 2, 3, 4\}$, output channels $\in \{16, 32, 64, 128\}$, kernel size $\in \{1, 3, 5\}$.

    \item Transformer: layers $\in \{1, 2, 3, 4\}$, number of attention heads $\in \{2, 4, 8, 16\}$, embedding size $\in \{32, 64, 128\}$.

    \item Informer: layers $\in \{1, 2, 3, 4\}$, hidden units $\in \{32, 64, 128\}$, number of attention heads $\in \{2, 4, 8, 16\}$, embedding size $\in \{32, 64, 128\}$.

    \item PureKAN: layers $\in \{1,2,3,4\}$
    
\end{itemize}
In addition to the above model-specific parameters, training parameters are considered -- batch size $\in \{16, 32, 64, 128\}$ and learning rate $\in \{0.001, 0.0001, 0.00001\}$. The best-performing configurations for each baseline are
\begin{itemize}
    
    \item MLP: 3 layers with 128 hidden units each; Batch size: 64; Learning rate: 0.001.

    \item LSTM: 2 layers with 128 hidden units each and 0.1 dropout; Batch size: 32; Learning rate: 0.0001.

    \item GRU: 3 layers with 128 hidden units each and 0.1 dropout; Batch size: 32; Learning rate: 0.0001.

    \item TCN: 4 layers with 128 output channels and kernel size of 3; Batch size: 64; Learning rate: 0.0001.

    \item Transformer: 4 layers with 8 attention heads and embedding size of 128; Batch size: 32; Learning rate: 0.00001.

    \item Informer: 3 layers with 8 attention heads and embedding size of 128; Batch size: 64; Learning rate: 0.00001.

    \item PureKAN: 2 layers; Batch size: 32; Learning rate: 0.001.
    
\end{itemize}

\section{Relationships between Raw Input Mobility and Corresponding Scalar Representation} \label{appendix:relation_mobility_representation}

\begin{figure}[!t]
    \centering
    \includegraphics[width=\linewidth]{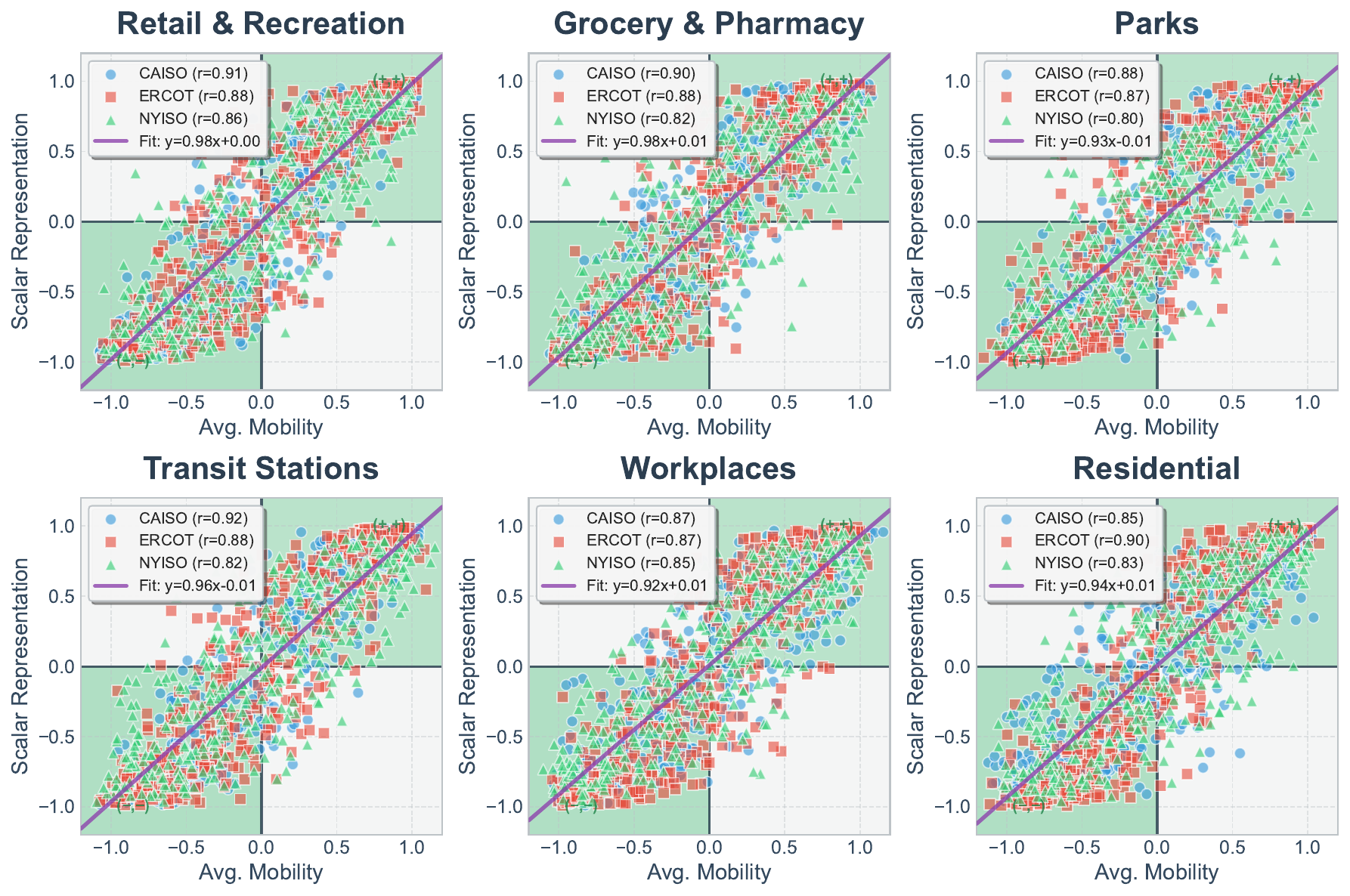}
    \caption{Relationships between average mobility and scalar representation.}
    \label{fig:relation_mobility_representation}
\end{figure}

To facilitate a direct comparison between the temporal mobility sequences and the resulting scalar representations, we calculate the average value of the mobility sequence as the primary metric for the x-axis in the plots in Fig. \ref{fig:relation_mobility_representation}. This averaging process enables a point-wise comparison between the raw input's magnitude and the scalar feature representation. Moreover, the average mobility  is an essential and physically intuitive metric to evaluate the historical mobility level, as it captures the net intensity of activity within the history window, making it a reasonable proxy for the total volume of movement that drives electricity demand.

The results of Fig. \ref{fig:relation_mobility_representation} provide an empirical proof that the scalar representation preserves both the scale and the sign of the physical input. As illustrated by the green-shaded regions, the data points predominantly reside in the $(+,+)$ and $(-,-)$ quadrants, suggesting that a physical increase in mobility consistently results in a positive representation, while a decline results in a negative one. Such strong and consistent positive correlations can be observed across all six mobility features and three markets. For example, in the Retail \& Recreation category, correlation coefficients (denoted as $r$) range from 0.86 to 0.91. Furthermore, the linear fit lines in blue across all features show slopes near unity with intercepts near zero. This near 1:1 mapping demonstrates that the scalar representation is not merely an abstract number but a high-fidelity proxy that respects the original feature's scale and sign.

Our framework is uniquely capable of capturing these clean, positive correlations due to the structural advantages of our feature-isolated attention architecture. In traditional MLP-based models, feature representations are often entangled through global matrix multiplications, making the resulting values a complex byproduct of interaction with other variables. Conversely, our model ensures that the representation for a specific mobility feature is a direct functional mapping of that indicator's own sequence alone. By combining this isolated processing with the local, univariate properties of KAN's B-spline activation functions, our LoadKAN is structurally incentivized to learn monotonic mappings between the raw input intensity and the final scalar representation. This transparency prevents the black-box mixing typical of deeper architectures and ensures that the sign and magnitude of the representation maintain a clear, interpretable link to the physical mobility dynamics they represent.

\section{Quantitative Feature Importance Analysis} \label{appendix:feature_importance_gradient_correlation}

We calculate feature importance scores (FIS) and correlations between learned spline gradients and load changes. For the former, instead of using Shapley values (which is a post-hoc approach, whereas our LoadKAN is inherently interpretable), we here calculate the L1-norm (i.e., the average absolute magnitude) of the learned spline activation functions. If the spline has a large amplitude, the corresponding feature contributes more to the load prediction. For the gradient-load correlation (GLC), we calculate the Pearson correlations between the derivative of the spline and the load changes.

\begin{table}[!h]
    \centering
    \caption{The calculated FIS in three electricity markets. Higher values indicate greater contribution to the forecast.}
    
    \resizebox{.9\columnwidth}{!}{%
    \begin{tabular}{l|c|c|c}
        \hline
        \textbf{Feature Category} & \textbf{NYISO} & \textbf{CAISO} & \textbf{ERCOT} \\ \hline
        Historical Load & 0.885 & 0.910 & 0.945 \\
        Weather (Temp \& Dew) & 0.720 & 0.765 & 0.810 \\ \hline
        Retail \& Recreation & 0.355 & \textbf{0.510} & 0.330 \\
        Grocery \& Pharmacy & 0.310 & 0.425 & 0.315 \\
        Parks & 0.580 & 0.290 & \textbf{0.612} \\
        Transit Stations & \textbf{0.445} & 0.380 & 0.210 \\
        Workplaces & 0.320 & \textbf{0.495} & 0.385 \\
        Residential & 0.410 & 0.360 & 0.395 \\ \hline
    \end{tabular}%
    \label{tab:FIS}
}
\end{table}

The FIS results are reported in Table \mbox{\ref{tab:FIS}}, which strongly corroborate the qualitative patterns observed in the activation visualizations of all three electricity markets.
\begin{itemize}
    
    \item In NYISO, the \textit{Transit Stations} mobility achieves a score of $0.445$, significantly higher than in ERCOT ($0.210$), reflecting the high dependence on public transport in New York and its utility as a proxy for urban energy activity.

    \item In CAISO, the mobility feature importance is more distributed among commercial indicators, with \textit{Retail \& Recreation} ($0.510$) and \textit{Workplaces} ($0.495$) showing consistently high contributions.

    \item In ERCOT, the \textit{Parks} feature exhibits an exceptionally high importance score of $0.612$, surpassing even typical commercial drivers like \textit{Workplaces} ($0.385$). This quantitatively confirms our finding that outdoor recreational behavior (and associated HVAC loads) is a dominant driver in the Texas market.
    
\end{itemize}

The results of GLC are reported in Table \mbox{\ref{tab:GLC}}, further elucidating the nature of these relationships. Notably, the \textit{Residential} feature consistently shows negative correlations (e.g., $-0.42$ in NYISO, $-0.38$ in CAISO). This quantitatively validates the \textit{substitution effect} hypothesis discussed in Section \mbox{\ref{subsubsec:exp_results_discussion}}: increased residential presence correlates with a net decrease in system-wide load, likely due to the concurrent reduction in higher-intensity commercial and industrial consumption. Furthermore, the strong positive correlation of \textit{Parks} in ERCOT ($0.68$) indicates that the grid load is highly sensitive and directionally aligned with fluctuations in park visitation, suggesting a direct link between outdoor activity conditions (likely weather-driven) and cooling demand.

\begin{table}[!t]
    \centering
    \caption{The calculated GLC in three electricity markets. Positive values indicate a pro-cyclic relationship; negative values indicate a counter-cyclic (substitution) relationship.}
    
    \resizebox{.9\columnwidth}{!}{%
    \begin{tabular}{l|c|c|c}
        \hline
        \textbf{Feature Category} & \textbf{NYISO} & \textbf{CAISO} & \textbf{ERCOT} \\ \hline
        Retail \& Recreation & 0.45 & 0.55 & 0.41 \\
        Grocery \& Pharmacy & 0.38 & 0.48 & 0.42 \\
        Parks & 0.52 & 0.15 & \textbf{0.68} \\
        Transit Stations & \textbf{-0.35} & 0.42 & 0.12 \\
        Workplaces & 0.28 & 0.45 & 0.31 \\
        Residential & \textbf{-0.42} & \textbf{-0.38} & \textbf{-0.25} \\ \hline
    \end{tabular}%
    }
    \label{tab:GLC}
    
\end{table}

\section*{Acknowledgement}
This work was supported in part by the Australian Research Council (ARC) Discovery Early Career Researcher Award (DECRA) under Grant DE230100046.

\bibliographystyle{elsarticle-num} 
\bibliography{ref}
\end{document}